\documentclass[acmsmall,screen]{acmart}

\acmJournal{TOSEM}

\setcopyright{none}
\acmDOI{}
\acmYear{}
\copyrightyear{}

\AtBeginDocument{%
  }

\newcommand{\up}{\textcolor{green}{$\blacktriangle$}}
\newcommand{\down}{\textcolor{red}{$\blacktriangledown$}}

\usepackage{multirow}

\usepackage{diagbox}
\usepackage{graphicx}
\usepackage{xcolor}
\usepackage{tcolorbox}
\usepackage[normalem]{ulem}
\usepackage{hyperref}

\usepackage{pgfplots}
\pgfplotsset{compat=1.18}
\usepgfplotslibrary{groupplots}
\pgfplotsset{compat=1.18}
\begin{document}

\title{Standing on the Shoulders of Giants: Stabilized Knowledge Distillation for Cross--Language Code Clone Detection}


\author{Mohamad Khajezade}
\affiliation{%
  \institution{University of British Columbia, Okanagan Campus}
  \city{Kelowna}
  \country{Canada}}
\email{khajezad@student.ubc.ca}

\author{Fatemeh H. Fard}
\affiliation{%
  \institution{University of British Columbia, Okanagan Campus}
  \city{Kelowna}
  \country{Canada}}
\email{fatemeh.fard@ubc.ca}

\author{Mohamed S. Shehata}
\affiliation{%
 \institution{University of British Columbia, Okanagan Campus}
 \city{Kelowna}
 \state{British Columbia}
 \country{Canada}}
\email{mohamed.sami.shehata@ubc.ca}




\renewcommand{\shortauthors}{Khajezade et al.}

\begin{abstract}
to be completed
\end{abstract}




\begin{abstract}
Cross-language code clone detection (X-CCD) is challenging because semantically equivalent programs written in different languages often share little surface similarity. Although large language models (LLMs) have shown promise for semantic clone detection, their use as black-box systems raises concerns about cost, reproducibility, privacy, and unreliable output formatting. In particular, compact open-source models often struggle to follow reasoning-oriented prompts and to produce outputs that can be consistently mapped to binary clone labels.

To address these limitations, we propose a knowledge distillation framework that transfers reasoning capabilities from DeepSeek-R1 into compact open-source student models for X-CCD. Using cross-language code pairs derived from Project CodeNet, we construct reasoning-oriented synthetic training data and fine-tune Phi3 and Qwen-Coder with LoRA adapters. We further introduce response stabilization methods, including forced conclusion prompting, a binary classification head, and a contrastive classification head, and evaluate model behavior using both predictive metrics and response rate. 

Experiments on Python--Java, Rust--Java, Rust--Python, and Rust--Ruby show that knowledge distillation consistently improves the reliability of compact models and often improves predictive performance, especially under distribution shift. In addition, classification-head variants substantially reduce inference time compared to generation-based inference. Overall, our results show that reasoning-oriented distillation combined with response stabilization makes compact open-source models more practical and reliable for X-CCD detection. 
\end{abstract}

\maketitle

\newpage

\section{Introduction}

Code Clone Detection (CCD) is a fundamental task in software engineering, as duplicated or near-duplicated code can lead to increased maintenance effort, bug propagation, and reduced software quality \cite{zhang2021survey}. Beyond maintenance, CCD is widely used in software forensics, code search, security analysis, and plagiarism detection \cite{du2024adaccd}. As modern software systems increasingly integrate multiple programming languages, the CCD problem extends beyond single-language settings to cross-language clone detection (X-CCD). In this setting, identifying clones often requires reasoning about program semantics and developer intent rather than relying on surface-level syntactic similarity \cite{alshabib2025systematic}.

Traditional clone detection approaches—such as text-based, token-based, syntactic, and metric-based techniques—perform well for Type-1 and Type-2 clones, where substantial syntactic overlap is preserved  \cite{alazba2024cort, mohammed2022search}. However, their performance degrades significantly for Type-3 and, in particular, Type-4 clones, where semantic equivalence must be inferred despite substantial structural differences \cite{lei2022deep, yu2019neural}. Deep learning-based methods partially address this limitation by learning code representations, yet prior studies show that these models struggle with out-of-distribution inputs and exhibit limited generalization in cross-language scenarios \cite{liu2021can, sonnekalb2022generalizability, tao2022c4}.

Large Language Models (LLMs) have recently demonstrated strong zero-shot performance on a wide range of code understanding tasks and have been explored for CCD. Prior work \cite{dou2023towards, moumoula2025struggles} shows that models such as GPT-3.5, GPT-4, LLaMA-2-Chat, and StarCoder can outperform classical CCD tools when semantic reasoning is required, and that chain-of-thought (CoT) prompting further improves semantic clone detection. However, these models are used strictly as black boxes: they remain unmodified, and their performance depends entirely on prompting strategies. This usage introduces several limitations:
(1) open-source models consistently underperform compared to proprietary models,
(2) smaller models are highly sensitive to prompt length and often fail to reliably follow long reasoning instructions, and
(3) API-based solutions raise concerns related to cost, privacy, and reproducibility.

A promising direction is knowledge distillation (KD). Recent work on knowledge distillation--most notably in the Microsoft Phi series \cite{gunasekar2023textbooks}--shows that compact models can acquire strong reasoning capabilities when trained on synthetic, ``textbook-style'' data generated by a larger teacher model. 
Knowledge distillation has been studied for different code understanding tasks, including code generation \cite{li2025symmetry}, code classification, code summarization \cite{d2025compression}, code search \cite{d2025compression}, and code clone detection \cite{wang2025empirical}. 
While KD has been investigated in the literature, it has mainly been applied to code understanding, with X-CCD treated as a downstream task. Our approach, however, is specific to CCD and addresses this problem exclusively. Additionally, existing studies have primarily focused on masked language models such as CodeBERT. To the best of our knowledge, no prior work has applied KD to autoregressive language models for reasoning, nor has it been explored for X-CCD.

These works motivate the central goal of this work: \textbf{to investigate whether reasoning capabilities relevant to cross-language CCD can be effectively distilled from a strong teacher model into compact open-source models suitable for local deployment, addressing the need for small open-source language models with high performance in cross-language CCD. }

In this paper, we develop a knowledge-distillation pipeline using DeepSeek-R1 as the teacher model and fine-tune two compact student models, Phi-3-Mini and Qwen2.5-Coder-3B. For the remainder of this paper, we refer to them as Phi-3 and Qwen-Coder, respectively. We generate System~2-style reasoning, which refers to a type of reasoning that mimics the slow, deliberate, and step-by-step thinking process humans use for complex problems \cite{yang2025llm2}, traces for cross-language clone pairs spanning Python-Java, Rust-Java, Rust-Python, and Rust-Ruby. Using these traces, we construct multiple variants of synthetic training datasets and analyze how different distillation formats affect final model performance. 
We evaluate the resulting models under two generalization settings: \emph{unseen programming languages} and \emph{unseen problems}. For unseen programming languages, the models are trained on Python-Java pairs and evaluated on Rust-Java, Rust-Python, and Rust-Ruby pairs, where test sets are drawn from the same problem distribution as the training set. For unseen problems, test sets are generated from a different problem distribution than that used for training.
An additional limitation in prior CCD studies using LLMs is that, in some cases, model outputs cannot be mapped to binary clone/non-clone labels. Most existing works treat such responses as failures when computing performance metrics \cite{moumoula2024large}. To address this issue, we introduce a new metric, \emph{response rate}, to more accurately capture model behavior. Furthermore, we propose two novel methods to achieve a complete response rate (100\%): \emph{forced conclusion} prompting and the addition of a classification head. For the latter, we explore two different training objectives: a simple binary classification head and a contrastive classifier. 

Accordingly, our study is guided by the following research questions:

\begin{enumerate}
\item \textbf{RQ1:} How well do small open-source models perform on X-CCD for unseen programming languages and unseen problems? This question investigates both detection performance and response rate.
\item \textbf{RQ2:} Can knowledge distillation from a reasoning-focused teacher model improve cross-language CCD performance and response rate of small student models?
\item \textbf{RQ3:} What is the most effective approach to achieving a complete response rate?
\end{enumerate}

Our results show that applying knowledge distillation to the baseline language models, Phi3 and Qwen-Coder, increases the response rate across all programming language pairs studied in this work, for both same-distribution (SD) and different-distribution (DD) problems. For instance, KD increases the response rate of Qwen-Coder from 24.3 to 67.7 on SD Rust--Ruby problems, and that of Phi3 from 34.6 to 53.8 on DD Python--Java problems. Additionally, KD improves the performance of our forced-conclusion method across all programming language pairs, for both SD and DD settings. For example, KD increases the F1 score of the forced-conclusion method by 10.21\% for SD Rust--Java problems using Phi3, and by 10.5\% for Qwen-Coder on DD problems. 
Moreover, while KD improves performance in 62\% of scenarios when adding a binary classification head and in 50\% of scenarios when adding a contrastive classification head, further analysis shows that incorporating a classification head can reduce inference time from hours to only a few minutes. 
These findings show that the proposed response-stabilization methods are essential for making LLM-based X-CCD practical: forced conclusion preserves the model's reasoning ability while guaranteeing a valid binary decision, the binary classification head provides a fast and deterministic alternative to generation-based inference, and the contrastive classification head encourages clone and non-clone pairs to form more separable representations under distribution shift.
This paper makes the following contributions:
\begin{enumerate}
    \item We create the first reasoning-oriented synthetic training dataset for cross-language CCD, consisting of 10{,}671 samples generated using DeepSeek-R1 and covering four programming language pairs. We use this dataset to evaluate generalization across unseen problems and unseen programming languages.
    \item We systematically analyze four dataset variants that differ in prompt structure and reasoning content, identifying which formats are most effective for distilling CCD-specific reasoning.
    \item We fine-tune two compact open-source models, Phi3 and Qwen-Coder, using LoRA and show that knowledge distillation improves CCD performance in both same-distribution and different-distribution settings. Specifically, the scaled average F1 score improves by 6.92 and 8.84 points for Phi3 in the SD and DD settings, respectively, and by 27.28 and 23.38 points for Qwen-Coder.
    \item We evaluate the cross-language generalization of distilled models, providing one of the first studies on multilingual CCD distillation.
    \item We introduce two novel methods to address incomplete response behavior in language models for X-CCD.
    \item We open-source all our datasets and trained models\footnote{\href{https://github.com/FARD-Lab/lm-xccd}{GitHub Repo}.}\footnote{\href{https://huggingface.co/mkhfring/lm-xccd}{Models Repo}.}.
\end{enumerate}

The rest of this paper is organized as follows. In Section \ref{bkg}, we review concepts related to our research. In Section \ref{std}, we describe the methodology used in our study, along with the datasets, baselines, and evaluation measures. Section \ref{experiments} presents the experiments conducted, and Section \ref{results} reports the results. Section~\ref{discussion-section} discusses some of the results in more detail. In Section \ref{ltr}, we review previous studies related to this work. Section \ref{validity} discusses threats to validity, and finally, Section \ref{conclusion} concludes the paper. 
\section{Background}\label{bkg}
This section reviews the main concepts relevant to our study: instruction-following language models and knowledge distillation for compact model adaptation.

\subsection{Instruction-Following Models}

Instruction-following models are language models trained or fine-tuned to generate responses based on natural language instructions. Rather than only continuing text, these models are optimized to follow user intent and produce task-specific outputs ~\cite{ouyang2022training}. This capability is especially important for tasks such as code clone detection, where the model must compare two code snippets, reason about their functionality, and produce a final decision.

Prior work has improved instruction-following behavior through supervised fine-tuning on instruction--response datasets. For example, FLAN demonstrates that instruction tuning can improve generalization to unseen tasks, while InstructGPT shows that aligning language models with human instructions leads to more useful and reliable responses~\cite{wei2021finetuned,ouyang2022training}. Open-source instruction-tuned models, such as Alpaca and later LLaMA-based variants, further demonstrate that compact or locally deployable models can be adapted to follow task-specific prompts ~\cite{taori2023alpaca,zhang2023llamaadapter}. However, full fine-tuning of large language models can be computationally expensive, motivating parameter-efficient adaptation methods ~\cite{hu2022lora}.

In this work, we use compact instruction-tuned autoregressive language models as student models for X-CCD. Because the task requires both semantic reasoning and reliable binary decisions, instruction-following ability is central to our formulation.

\subsection{Knowledge Distillation}

Knowledge distillation is a model compression technique that transfers knowledge from a large teacher model to a smaller student model. Bucil{\u{a}} et al.~\cite{bucila2006model} first introduced the idea of model compression, where a compact model is trained to approximate the behavior of a larger model or an ensemble while maintaining reasonable accuracy. This idea was later formalized as \emph{knowledge distillation} by Hinton et al.~\cite{hinton2015distilling}. In a typical knowledge distillation setting, the teacher model provides supervision signals, such as soft labels, explanations, or generated outputs, and the student model is trained to reproduce the teacher's behavior.

In the context of language models, knowledge distillation can be used not only to transfer final predictions but also to transfer reasoning patterns. This is particularly useful for X-CCD, where two programs written in different languages may implement the same functionality despite having different syntax and structure. A strong teacher model can generate reasoning traces that explain functional similarity, algorithmic behavior, and structural differences between code snippets. A smaller student model can then be fine-tuned on these teacher-generated traces to improve its ability to perform semantic clone detection.

In this work, we use DeepSeek-R1 ~\cite{guo2025deepseekr1} as the teacher model and fine-tune compact open-source autoregressive student models using teacher-generated reasoning and conclusion outputs. To make this adaptation efficient, we use Low-Rank Adaptation (LoRA), a parameter-efficient fine-tuning method that introduces trainable low-rank matrices into selected layers of the model while keeping most of the original model parameters frozen. This allows the student models to acquire task-specific X-CCD reasoning behavior without requiring full model fine-tuning.

A knowledge distillation system generally consists of three components: the knowledge being transferred, the distillation algorithm, and the teacher--student relationship. In our study, the transferred knowledge consists of reasoning-oriented explanations and clone/non-clone conclusions; the distillation algorithm is LoRA-based fine-tuning on the curated synthetic dataset; and the teacher--student relationship is defined by using DeepSeek-R1 to supervise compact student models, namely Phi3 ~\cite{abdin2024phi3} and Qwen-Coder ~\cite{hui2024qwen25coder}.

\section{Study Design}\label{std}
\begin{figure}[H]
    \centering
    \includegraphics[width=\textwidth]{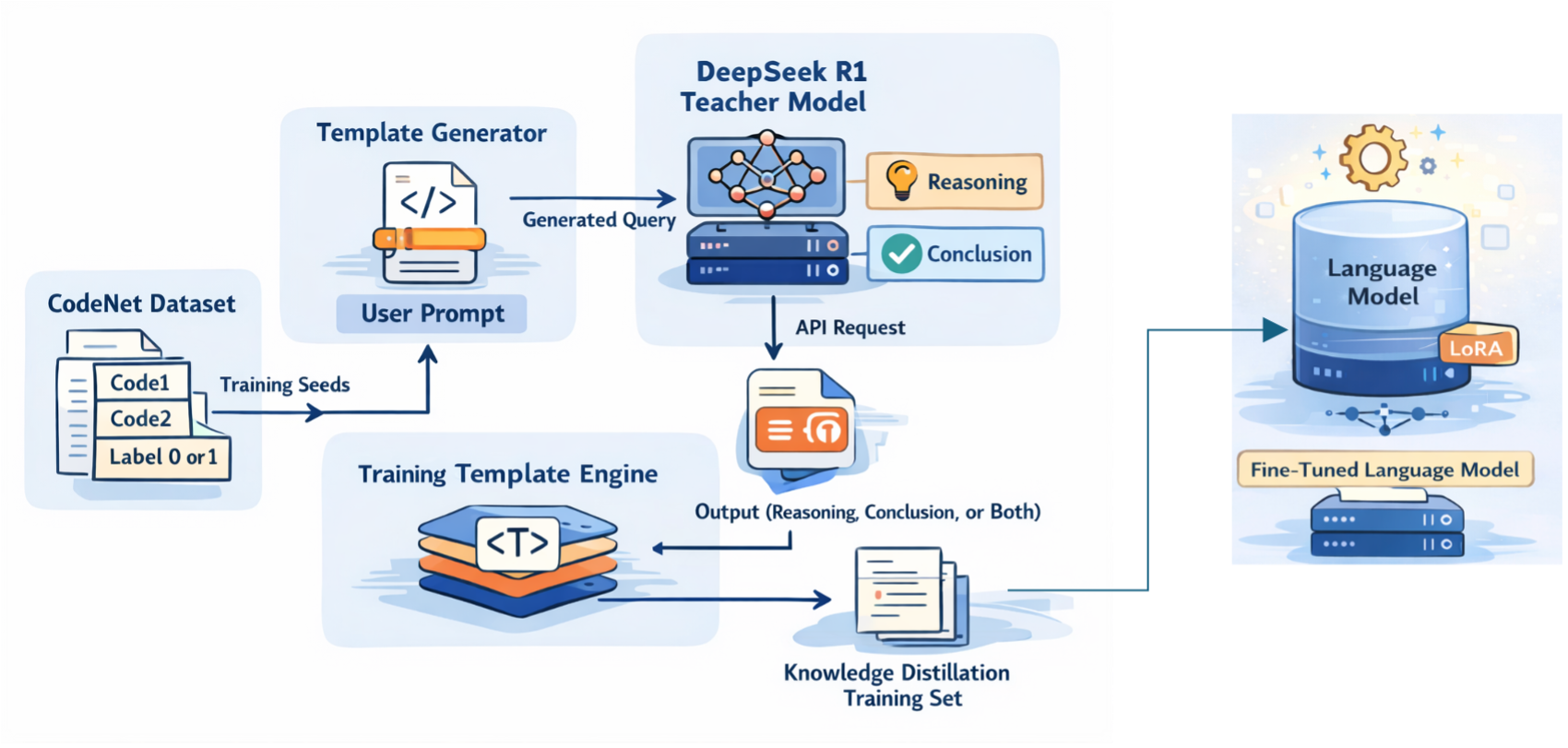}
    \caption{ \textbf{Knowledge distillation and fine-tuning methodology:} A seed dataset is generated from the Project CodeNet dataset, where each sample consists of code1, code2, and a label indicating whether the pair is a clone (1) or non-clone (0). A template generator embeds each sample into a user prompt used to query the DeepSeek R1 teacher model. The teacher output includes reasoning and conclusion components, which are processed by a template engine to create multiple knowledge distillation training sets using two user prompt variations and three system prompt configurations (reasoning, conclusion, or both). The resulting training sets are used to fine-tune small language models (Phi3 and Qwen-Coder) using LoRA adapters for X-CCD.} 
    \Description{} 
    \label{fig:distillation_pipeline}
\end{figure}

In this section, we present the methodology used to design this study. We first elaborate on our approach to knowledge distillation, which is based on the method proposed in~\cite{haider2024phi}. Accordingly, we begin by curating various versions of knowledge distillation training dataset, as discussed in Section~\ref{data-construction}.
Each training dataset variant is used to train a separate student model, where the best version is chosen for our experiments. 
We then describe the knowledge distillation pipeline using this curated training set in Section~\ref{knowlege-pipe}. The overall approach is depicted in Figure \ref{fig:distillation_pipeline}. 

To stabilize the responses of our language models and obtain consistent binary labels, we introduce three methods: forced conclusion, binary classification head, and contrastive classification head, with details discussed in Section~\ref{response-stable}.

\subsection{Problem Definition}
Our study addresses X-CCD, which aims to determine whether two code snippets written in different programming languages implement the same functionality.
Given a pair of code snippets $x = (c_1, c_2)$, where $c_1$ and $c_2$ are written in different languages, the goal is to assign a binary label:
\begin{equation}
y \in \{0,1\},
\end{equation}
where $y = 1$ indicates that the two snippets are functionally equivalent (clone), and $y = 0$ otherwise.

Unlike traditional classification approaches, this work adopts a generation-based formulation \cite{khajezade2024investigating}. Instead of directly predicting a binary label, a language model $M$ is prompted with the input pair $x$ and generates a free-form natural language response:
\begin{equation}
r = M(x),
\end{equation}
which includes reasoning about the relationship between the two code snippets.

To obtain a binary prediction, the generated output $r$ is post-processed using a deterministic parsing function:
\begin{equation}
\hat{y} = g(r),
\end{equation}
where $g(\cdot)$ maps the textual response to a binary label using pattern matching (e.g., regular expressions).

However, due to the open-ended nature of language model outputs, the response $r$ is not always directly interpretable as a valid binary decision. This leads to ambiguity and reduces the reliability of evaluation.

To address this limitation, we later introduce response stabilization methods (Section~\ref{response-stable}) that enforce a consistent mapping from model outputs to binary labels. These methods transform the inherently generative outputs of language models into reliable binary predictions, enabling consistent evaluation for X-CCD.

\subsection{Dataset Construction}\label{data-construction}
Knowledge distillation frequently relies on generating synthetic training data using a strong teacher model. Moreover, previous work indicates that chain-of-thought reasoning can significantly improve the accuracy of language models~\cite{wang-etal-2023-towards}. Following prior work~\cite{gunasekar2023textbooks}, we leverage a teacher model to synthesize high-quality, reasoning-oriented X-CCD data.

Effective training of a reasoning-based model requires that the data exhibit System~2-style analytical reasoning~\cite{xiang2025towards}, i.e., deliberate, step-by-step logical inference. Consequently, the prompts used in this study explicitly instruct the teacher model to produce System~2-aligned chain-of-thought explanations when determining whether two code fragments constitute a clone pair. This subsection describes the process of seed data generation and the use of a teacher model to construct a knowledge-distilled training dataset for X-CCD.

\subsubsection{Seed Data from CodeNet}\label{dataseed}

To construct a high-quality seed dataset for X-CCD, we leverage the Project CodeNet dataset \cite{puri2021codenet}, which contains a large collection of competitive programming solutions written in multiple programming languages. CodeNet is particularly suitable for X-CCD because each problem is associated with multiple implementations that are functionally equivalent but syntactically diverse written in different languages.

Following prior work on X-CCD ~\cite{khajezade2024investigating}, we first filter the dataset to retain only correct submissions. Let $D_{\text{CodeNet}}$ denote the full set of submissions. We construct a filtered set:
\begin{equation}
D_{\text{valid}} = \{(s, p, \ell) \in D_{\text{CodeNet}} \mid \text{status}(s) = \text{Accepted}\},
\end{equation}
where $s$ is a submission, $p$ is the problem identifier, and $\ell$ is the programming language. This filtering ensures that all selected programs are functionally correct, reducing noise in the resulting dataset.

From $D_{\text{valid}}$, we construct cross-language code pairs. Positive (clone) pairs are created by selecting two submissions that solve the same problem but are written in different programming languages. Negative (non-clone) pairs are formed by sampling submissions from different problems, ensuring functional non-equivalence.

We focus on four programming languages: Python, Java, Rust, and Ruby. Python and Java are widely studied in prior literature \cite{moumoula2025struggles, yahya2022cross}, while Rust and Ruby are comparatively less explored, allowing us to evaluate performance across both high-resource and low-resource language settings.

Based on these languages, we construct a balanced seed training set, in which 50\% of the samples are positive and 50\% are negative, consisting of four cross-language subsets: Python--Java, Rust--Python, Rust--Java, and Rust--Ruby. The Python--Java subset contains 10{,}000 samples, while each of the remaining language pairs contains 2{,}000 samples. The larger size of the Python--Java subset reflects its prominence in prior work and serves as the primary training configuration, while the smaller subsets are included to extend coverage across additional language pairs under practical constraints on data generation cost.

Each sample is represented as a tuple $(c_1, c_2, y)$, where $c_1$ and $c_2$ are code snippets written in different languages, and $y \in \{0,1\}$ indicates whether the pair is a clone.

\subsubsection{Teacher-Guided Labeling with DeepSeek-R1}\label{teacher-labeling}

Motivated by the strong performance of DeepSeek-R1 on code-related tasks \cite{guo2025deepseek}, we leverage it as a teacher model to improve the quality and semantic richness of the training data. As illustrated in Figure~\ref{fig:distillation_pipeline}, each code pair $(c_1, c_2)$ from the seed curated in Section~\ref{dataseed} is embedded into a structured prompt using a template generator and submitted to the teacher model via an API call.

For each input pair, we construct a reasoning prompt, a structured instruction designed to elicit step-by-step reasoning prior to producing a final decision.

The reasoning prompt explicitly instructs the teacher model to analyze the code pair through the following steps:
\begin{itemize}
    \item \textbf{Functionality comparison:} Summarize each snippet and compare their functional behavior.
    \item \textbf{Algorithmic reasoning:} Analyze the underlying computational or mathematical processes.
    \item \textbf{Structural analysis:} Compare syntactic and structural properties of the implementations.
    \item \textbf{Similarity assessment:} Evaluate overall semantic similarity.
    \item \textbf{Conclusion:} Provide a final determination of clone or non-clone status.
\end{itemize}

\begin{figure}[t]
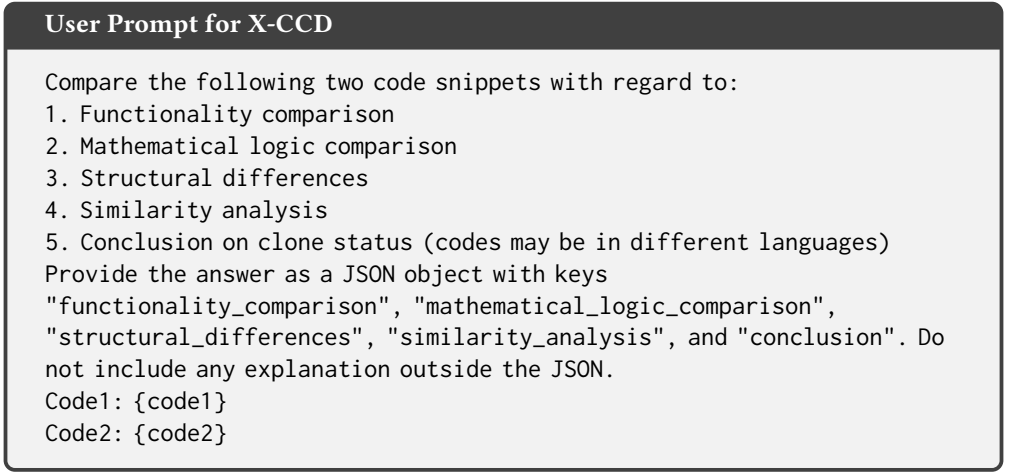


\centering
\begin{tcolorbox}[width=0.95\linewidth,
title=\textbf{User Prompt for X-CCD}, halign=left]
\ttfamily
Compare the following two code snippets with regard to:\\

1. Functionality comparison  \\
2. Mathematical logic comparison \\ 
3. Structural differences  \\
4. Similarity analysis  \\
5. Conclusion on clone status (codes may be in different languages)\\

Provide the answer as a JSON object with keys 
"functionality\_comparison", "mathematical\_logic\_comparison", 
"structural\_differences", "similarity\_analysis", and "conclusion". 
Do not include any explanation outside the JSON.

Code1: \{code1\}

Code2: \{code2\}
\end{tcolorbox}
\caption{This figure shows the actual user prompt used to encourage DeepSeek-R1 (the teacher model) to perform reasoning and conduct X-CCD.}
\Description{} 
\label{fig:user-prompt}

\end{figure}

Our prompt is represented in Figure~\ref{fig:user-prompt}.

The teacher model produces outputs that include both a reasoning component and a final conclusion. This enables the dataset to capture deeper semantic relationships between code snippets, which is essential for X-CCD.

\subsubsection{Training Data Curation and Augmentation}\label{teacher-trainset}

Let $\mathcal{D}_0 = \{(x_i, y_i)\}_{i=1}^N$ denote the set of seed X-CCD pairs, where each
\begin{equation}
x_i = \big(c_i^{(\ell_a)}, \; c_i^{(\ell_b)}\big)
\end{equation}
is a cross-language code pair and $y_i \in \{0,1\}$ is the corresponding ground-truth label.

For each input $x_i$, the teacher model produces two outputs: 

\begin{equation}
R_i = T(x_i)_{\text{reasoning}}, \qquad
C_i = T(x_i)_{\text{conclusion}},
\end{equation}

where $R_i$ is the reasoning part of the teacher output, and $C_i$ is the textual output of DeepSeek-R1, which includes the conclusion for the X-CCD task. To ensure label reliability, we retain only samples where the teacher's predicted label agrees with the ground truth:
\begin{equation}
\delta_i =
\begin{cases}
1, & \text{if } C_i^{(\text{label})} = y_i, \\
0, & \text{otherwise}.
\end{cases}
\end{equation}

Only samples with $\delta_i = 1$ are included in the final dataset. This filtering step reduces noise and prevents incorrect supervision signals from propagating to the student model.

To enrich the training data, we construct multiple dataset variants by combining prompt types with different components of the teacher outputs. To this end, we introduce an additional user prompt, referred to as the simple prompt, which asks only whether the provided pair of code snippets constitutes a code clone. This simple prompt is used solely to construct the training dataset and is not used to query the teacher model. The main motivation for using this simple prompt is to evaluate whether we can train a model that performs well with a simple prompt rather than a more complex prompt, such as the one shown in Figure~\ref{fig:user-prompt}. 
Let
\begin{equation}
\mathcal{V} = \{SR, SC, RR, RC, RRC\}
\end{equation}
denote the set of variants:
\begin{equation}
\begin{aligned}
Simple_{up}-Reasoining_{mr}(SR)  &= (\text{simple prompt}, \oplus R_i), \\
Simple_{up}-Conclusion_{mr}(SC) &= (\text{simple prompt}, \oplus C_i), \\
Reasoning_{up}- Reasoining_{mr} (RR) &= (\text{reasoning prompt}, \oplus R_i), \\
Reasoning_{up}- Conclusion_{mr} (RC) &= (\text{reasoning prompt}, \oplus C_i), \\
Reasoning_{up}-Reasoning + Conclusion_{mr} (RRC) &= (\text{reasoning prompt}, \oplus R_i \oplus C_i),
\end{aligned}
\end{equation}
where $\oplus$ denotes concatenation, $up$ represents the user prompt, and $mr$ represents the model response.
These variants provide different levels of supervision, ranging from direct classification signals to detailed reasoning-based explanations. We aim to investigate which part of the teacher output—reasoning, conclusion, or a combination of both—is most useful for KD. As a result, the first two variants, SR and SC, are designed to train a model that can operate with a simple prompt rather than requiring a more elaborate X-CCD-specific prompt. The other three variants continue to encourage the model to perform reasoning based on various criteria. In essence, the first two variants aim to reduce the need for prompt engineering in X-CCD, whereas the last three variants still train a model that remains sensitive to prompt design. That said, our preliminary experiments, as reported in Table~\ref{tab:original_results}, indicate that the first two variants are not effective. Moreover, the combination of reasoning and conclusion yields the best KD model. As a result, we use the RRC variant in the remainder of the experiments reported in this paper.

\subsection{Knowledge Distillation Pipeline}\label{knowlege-pipe}

Building on the curated dataset, we construct a knowledge distillation (pipeline to transfer reasoning capabilities from the teacher model to smaller student models (Figure~\ref{fig:distillation_pipeline}).
Each dataset variant $v_k$ defines a corresponding training dataset $\mathcal{D}_k$, and a separate student model $S^{(k)}_\theta$ is fine-tuned on each dataset.
To enable efficient adaptation, we employ Low-Rank Adaptation (LoRA), which introduces trainable low-rank matrices while keeping the base model largely frozen.

Given a tokenized input $(\mathbf{x}, \mathbf{m})$, the student model produces hidden representations:
\begin{equation}
\mathbf{H} = \left(\mathbf{h}_1, \mathbf{h}_2, \dots, \mathbf{h}_T\right), \quad \mathbf{h}_t \in \mathbb{R}^d.
\end{equation}

A fixed-size representation is obtained via attention-mask-aware mean pooling:
\begin{equation}
\mathbf{h}_{\text{pool}} =
\frac{\sum_{t=1}^{T} \mathbf{h}_t \cdot m_t}{\sum_{t=1}^{T} m_t}.
\end{equation}

Each student model is trained using a next-token prediction objective to imitate the teacher outputs:
\begin{equation}
\theta_k^\ast =
\arg\min_{\theta}
\sum_{z \in \mathcal{D}_k}
\mathcal{L}_{\mathrm{LM}}(S^{(k)}_\theta, z),
\end{equation}
where the language modeling loss is:
\begin{equation}
\mathcal{L}_{\mathrm{LM}}(S^{(k)}_\theta, z)
=
-\sum_{t=1}^{|z|}
\log p_\theta(z_t \mid z_{<t}).
\end{equation}

This formulation enables the student models to learn both classification behavior (via conclusions) and reasoning patterns (via chain-of-thought supervision), resulting in improved semantic understanding for X-CCD.

\subsection{Response Stabilization Methods}\label{response-stable}

While the knowledge distillation pipeline improves the model’s ability to perform code clone detection, we observed that the reliability of model outputs remains a critical challenge. In particular, large language models may generate responses that are ambiguous, verbose, or not directly mappable to binary labels, which negatively impacts evaluation metrics such as F1 score and response rate.

To address this issue, we introduce a set of response stabilization methods that aim to enforce deterministic and interpretable outputs. These methods are applied after the knowledge distillation step and are designed to reduce ambiguity while preserving the model’s semantic understanding. Specifically, we investigate three approaches: forced conclusion, binary classification head, and contrastive classification head.

\subsubsection{Forced Conclusion}

One of the main challenges in applying language models to binary classification tasks, such as code clone detection, is mapping the model’s generative output to a discrete binary label. This challenge is amplified when the model is encouraged to perform explicit reasoning, since prompting it to respond directly with a constrained output such as “yes” or “no” can limit its ability to express intermediate analysis.

To address this issue, we propose a two-stage inference procedure that enables unconstrained reasoning while ensuring a deterministic mapping to a binary decision. In the first stage, given an input sequence $x$, we prompt a language model $M_\theta$ to generate a free-form response for X-CCD, $r = M_\theta(x)$, which captures the model’s reasoning about the task. This response is not restricted in length or structure and may contain natural-language explanations or justifications. The prompts used in this stage are described in Section~\ref{teacher-trainset}.

In the second stage, we construct a follow-up prompt $x^{'} = g(x, r)$ that appends the original input and the generated response $r$ with an instruction asking the model to analyze its prior output and produce a binary judgment. The second prompt is illustraded in Figure \ref{fig:second-prompt}.

\begin{figure}[t]
\centering
\begin{tcolorbox}[
    width=0.95\linewidth,
    title=\textbf{Forced conclusion second stage prompt},
    halign=left
]
\ttfamily
Based on the following analysis, determine the final clone conclusion.\\
- You must decide whether the two codes are clones.\\
- Do not repeat the analysis. Just think internally and decide.\\
- We will infer your answer from the probabilities of the next token.\\[1em]

\{full\_response\}\\[1em]

- Final Answer (Yes or No):
\end{tcolorbox}
\caption{This figure represents the prompt that we used in the second stage of our forced-conclusion. the $\{full\_response\}$ is the response of the model in the first stage }
\label{fig:second-prompt}
\end{figure}

Rather than allowing the model to generate a textual answer, we extract the probability distribution over the vocabulary induced by the final hidden-state embedding of the model. Let $\mathbf{h}_{\text{last}} \in \mathbb{R}^d$ denote the last-layer embedding corresponding to the decision position. The probability of any token t $\in \mathcal{V}$  is given by: 
\begin{equation}\label{eq:prbdist}
    P(t \mid x') = \text{softmax}(W \mathbf{h}_{\text{last}} + \mathbf{b})_t,
\end{equation}

where W and $\mathbf{b}$ are the output projection parameters.

Let $t_{\text{yes}}$ and $t_{\text{no}}$ denote the vocabulary tokens corresponding to the labels “yes” and “no”, respectively. The final binary prediction $\hat{y} \in \{0,1\}$ is obtained by directly comparing their probabilities based on equation \ref{eq:prblabels}:

\begin{equation} \label{eq:prblabels}
    \hat{y} =
\begin{cases}
1, & \text{if } P(t_{\text{yes}} \mid x') > P(t_{\text{no}} \mid x') \\
0, & \text{otherwise}.
\end{cases}
\end{equation}

Crucially, no tokens are sampled or decoded during the second stage; the decision is made solely based on the model’s token-level probability distribution. This design ensures that every input is mapped to a binary label while preserving the model’s capacity for reasoning in the first stage. This approach relies on the assumption that the second-stage prompt sufficiently constrains the output space to the two target tokens, allowing a reliable comparison between their probabilities. We further investigate other methods to achieve a perfect response rate. These methods are explained in section \ref{binary-classifier-head} and \ref{contrastive-learning}. 

\subsubsection{Binary Classification Head}\label{binary-classifier-head}

In addition to prompt-based strategies for extracting binary decisions from language models, we propose a classification-based approach in which a dedicated binary classifier head is attached to the final layer of a pretrained language model. This design removes the need to rely on constrained generation or token-level probability comparisons and instead directly learns a mapping from the model’s internal representations to binary labels.
Let $M_\theta$ denote a pretrained causal language model with parameters $\theta$. Given an input prompt $x$, constructed from a pair of code snippets, the model produces a sequence of hidden states $\mathbf{H} = \left( \mathbf{h}_1, \mathbf{h}_2, \dots, \mathbf{h}_T \right),$
where $\mathbf{h}_t \in \mathbb{R}^d$ is the hidden-state embedding at token position $t$ from the final transformer layer. To obtain a fixed-size representation for classification, we apply attention-mask-aware mean pooling over the final-layer hidden states following equation \ref{bclprojection}:
\begin{equation}\label{bclprojection}
    \mathbf{h}_{\text{pool}} =
\frac{\sum_{t=1}^{T} \mathbf{h}_t \cdot m_t}{\sum_{t=1}^{T} m_t},
\end{equation}

where $m_t \in \{0,1\}$ is the attention mask indicating valid (non-padding) tokens. This pooling strategy captures global semantic information from the entire prompt while remaining invariant to padding length. On top of the pooled representation, we introduce a lightweight binary classification head $f_\phi$, parameterized by $\phi$, consisting of a projection layer, a non-linear activation, dropout, and a final linear classifier as formulated in equation \ref{bclobjective}:
\begin{equation}\label{bclobjective}
    \mathbf{z} = \tanh(W_p \mathbf{h}_{\text{pool}} + \mathbf{b}_p),
\ell = W_c \mathbf{z} + b_c,
\end{equation}

where $W_p \in \mathbb{R}^{d \times d}, W_c \in \mathbb{R}^{1 \times d},$ and $\ell \in \mathbb{R}$ is the scalar logit corresponding to the binary decision.

During training, the base language model $M_\theta$ is kept frozen, and only the classifier head parameters $\phi = \{W_p, \mathbf{b}_p, W_c, b_c\}$ are optimized. This design choice reduces computational cost, prevents catastrophic forgetting, and ensures that the classifier learns to leverage the pretrained representations rather than altering them. The model is trained using a binary cross-entropy loss based on equation \ref{bclloss}:
\begin{equation}\label{bclloss}
    \mathcal{L} = - \left[ y \log \sigma(\ell) + (1 - y) \log (1 - \sigma(\ell)) \right],
\end{equation}
where $y \in \{0,1\}$ is the ground-truth label and $\sigma(\cdot)$ denotes the sigmoid function.
At inference time, given a prompt x, the model computes the predicted probability
$p = \sigma(\ell)$, which represents the likelihood that the input pair corresponds to a code clone. The final binary prediction $\hat{y}$ is obtained via thresholding formulated in equation \ref{bclresult}:

\begin{equation}\label{bclresult}
    \hat{y} =
\begin{cases}
1 & \text{if } p \geq 0.5 \\
0 & \text{otherwise}.
\end{cases}
\end{equation}

This approach provides a direct and deterministic mapping from language model representations to binary labels, eliminating reliance on prompt engineering or constrained decoding. By decoupling reasoning from decision-making, the language model is free to encode rich semantic and structural information about the code pair, while the classifier head learns an explicit decision boundary tailored to the clone detection task.

\subsubsection{Contrastive Classification Head}\label{contrastive-learning}

In addition to the binary head, we introduce a contrastive classification head to improve robustness, particularly under distribution shift. The motivation for this approach is to encourage the model to learn similarity-preserving representations, where clone pairs are mapped closer together in the embedding space and non-clone pairs are pushed apart.

Building on the same pooled representation $\mathbf{h}_{\text{pool}}$, the contrastive head is trained using a contrastive objective that captures pairwise relationships between samples. This reduces reliance on a single decision boundary learned from the training distribution and instead promotes more stable representations that generalize better to unseen samples.

Concretely, we reuse the same feature extractor and projection architecture as in the binary classification head. Given an input prompt $x$, the pretrained causal language model $M_\theta$ produces final-layer hidden states
\begin{equation}
\mathbf{H} = \left( \mathbf{h}_1, \mathbf{h}_2, \dots, \mathbf{h}_T \right),
\end{equation}
from which a fixed-size representation is obtained by attention-mask-aware mean pooling:
\begin{equation}
\mathbf{h}_{\text{pool}} =
\frac{\sum_{t=1}^{T} \mathbf{h}_t \cdot m_t}{\sum_{t=1}^{T} m_t}.
\end{equation}

The pooled representation is then mapped into a task-specific embedding space using a trainable projection head:
\begin{equation}\label{eq:contrastive_proj}
\mathbf{z} = \tanh(W_p \mathbf{h}_{\text{pool}} + \mathbf{b}_p),
\end{equation}
where $W_p \in \mathbb{R}^{d \times d}$ and $\mathbf{z} \in \mathbb{R}^{d}$. As in the binary classification head, dropout is applied after the projection layer, and a linear classifier is used to produce a scalar logit:
\begin{equation}\label{eq:contrastive_logit}
\ell = W_c \mathbf{z} + b_c.
\end{equation}

During training, the base language model $M_\theta$ is kept frozen, and only the parameters of the projection and classification head are optimized. Let a mini-batch contain $N$ samples with projected embeddings $\{\mathbf{z}_i\}_{i=1}^{N}$ and binary labels $\{y_i\}_{i=1}^{N}$, where $y_i \in \{0,1\}$. We first $\ell_2$-normalize the projected embeddings:
\begin{equation}
\tilde{\mathbf{z}}_i = \frac{\mathbf{z}_i}{\|\mathbf{z}_i\|_2}.
\end{equation}

We then compute pairwise similarities using scaled cosine similarity:
\begin{equation}
s_{ij} = \frac{\tilde{\mathbf{z}}_i^\top \tilde{\mathbf{z}}_j}{\tau},
\end{equation}
where $\tau > 0$ is a temperature hyperparameter. For each anchor sample $i$, samples in the same mini-batch with the same label are treated as positives, while samples with different labels are treated as negatives. Self-comparisons are excluded. Let
\begin{equation}
\mathbb{P}(i) = \{j \neq i \mid y_j = y_i\}
\end{equation}
denote the set of positive indices for anchor $i$. The supervised contrastive loss is defined as:
\begin{equation}\label{eq:supcon_loss}
\mathcal{L}_{\text{con}} =
-\frac{1}{N} \sum_{i=1}^{N}
\frac{1}{|\mathbb{P}(i)|}
\sum_{j \in \mathbb{P}(i)}
\log
\frac{\exp(s_{ij})}
{\sum_{k \neq i} \exp(s_{ik})}.
\end{equation}

This objective encourages embeddings of clone pairs to become more similar, while pushing embeddings of non-clone pairs farther apart. In practice, the contrastive head is trained jointly with the binary classification objective. Specifically, we combine the binary cross-entropy loss with the contrastive loss:
\begin{equation}\label{eq:joint_contrastive_loss}
\mathcal{L}_{\text{total}} =
\mathcal{L}_{\text{BCE}} + \lambda \mathcal{L}_{\text{con}},
\end{equation}
where $\mathcal{L}_{\text{BCE}}$ is the binary cross-entropy loss defined in Equation~\ref{bclloss}, and $\lambda$ is a weighting coefficient controlling the contribution of the contrastive term.

Thus, unlike a purely metric-learning formulation, our contrastive classification head retains an explicit classifier while simultaneously regularizing the embedding space. This design allows the model to benefit from both a direct decision boundary and a similarity-preserving representation.

At inference time, prediction is performed through the classifier head rather than nearest-neighbor retrieval. Given an input prompt $x$, the model computes the scalar logit $\ell$, converts it to a probability $p = \sigma(\ell)$, and produces the final binary prediction using the same thresholding rule as in the binary classification head:
\begin{equation}
\hat{y} =
\begin{cases}
1 & \text{if } p \geq 0.5 \\
0 & \text{otherwise}.
\end{cases}
\end{equation}

While prior work suggests that contrastive objectives can improve robustness \cite{zubkov2022evaluation}, the effectiveness of this approach depends on the design of the contrastive loss and the quality of the training data. In our setting, the contrastive head serves as an additional mechanism to stabilize predictions, complementing both the binary head and the forced conclusion approach.

\subsection{Inference Pipeline}

After fine-tuning, the trained student models are evaluated on unseen code pairs using a consistent inference pipeline. Given a pair of code snippets $(c_1, c_2)$, a structured prompt is constructed using the same template format employed during training to ensure alignment between training and inference distributions.

The inference procedure varies depending on the response stabilization method applied:

\begin{itemize}
    \item \textbf{Base Model:} The model generates a free-form natural language response. A binary label is extracted by applying regular expressions to map the output to either 0 or 1. Outputs that cannot be parsed are treated as invalid.

    \item \textbf{Forced Conclusion:} In this setting, inference is performed in two stages. First, the model generates a reasoning-based response $r = M(x)$ for the input pair $x$. In the second stage, a follow-up prompt $x' = g(x, r)$ is constructed and passed to the model. Instead of generating a textual response, the model’s output distribution is used to compute the probabilities of the target tokens corresponding to ``yes'' and ``no''. The final prediction is obtained by selecting the label with the higher probability (equation \ref{eq:prblabels}). 
    This approach eliminates the need for output parsing and ensures that every input yields a valid binary prediction.

    \item \textbf{Binary Classification Head:} The model directly outputs a scalar logit, which is converted to a probability using a sigmoid function. The final prediction is obtained via equation \ref{bclresult}. This method provides a deterministic mapping without requiring text generation. It is worth noting that this method was applied to both the baseline model and the KD model. In both cases, the layers of the language model were frozen, and a binary classifier was trained on top of the LLM using the ground-truth labels from the training data. During inference, the trained binary classifier outputs either 0 or 1.

    \item \textbf{Contrastive Classification Head:} The model produces embedding representations for the input pair, and the prediction is derived based on similarity in the learned embedding space. A threshold over similarity scores is used to determine the final binary label, ensuring consistent and interpretable outputs. Similar to the binary classification head, this method is applied to both the baseline model and the KD model. For the contrastive learning head, the joint objective in Equation~\ref{eq:joint_contrastive_loss} is used during training, while the thresholding rule in Equation~\ref{bclresult} is used during inference to map the model output to either 0 or 1.
    
\end{itemize}

To ensure fair comparison across methods, all models are evaluated on the same test sets under identical conditions. Predictions are compared against ground-truth labels to compute evaluation metrics.

For methods that rely on natural language outputs (i.e., the base model), cases where the output cannot be mapped to a valid label are treated as invalid predictions or accounted for separately using the response rate metric, as described in Section~\ref{metric-sub}.

\section{Experiments}\label{experiments}

This section describes the experimental setup used to evaluate the proposed knowledge-distillation and response-stabilization methods. The experiments were conducted in two stages. Table~\ref{tab:experiment_overview} summarizes the two experimental stages. First, we performed preliminary experiments on Python--Java code pairs to compare the five knowledge-distillation training variants introduced in Section~\ref{teacher-trainset}. 
The preliminary stage was designed to identify the most effective KD training format. 
Second, based on the preliminary results, we selected the best-performing variant and extended the evaluation to additional programming language pairs and response-stabilization methods.
The extended stage was designed to evaluate whether the selected KD setting generalizes across additional language pairs and whether response-stabilization methods improve the reliability of binary prediction.

\subsection{Models and Baselines}\label{baseline-sub}

We evaluate two compact instruction-tuned causal language models: Phi-3-mini-128k-instruct, denoted as Phi3, and Qwen2.5-Coder-3B-Instruct, denoted as Qwen-Coder. These models are used in two roles. First, they serve as baseline models, where the original pretrained models are evaluated without task-specific KD fine-tuning. Second, they serve as student models in our KD pipeline, where they are fine-tuned using teacher-generated responses from DeepSeek-R1.

This setup allows us to compare the original zero-shot capabilities of each model against its distilled version while controlling for model architecture. In addition, for the response-stabilization experiments, we evaluate classification heads on top of both the original baseline models and the KD fine-tuned models.


\begin{table}[t]
\centering
\caption{Overview of the experimental stages.}
\label{tab:experiment_overview}
\begin{tabular}{p{0.15\linewidth} p{0.31\linewidth} p{0.18\linewidth} p{0.26\linewidth}}
\toprule
\textbf{Stage} & \textbf{Language pairs} & \textbf{KD variants} & \textbf{Inference and Stabilization methods} \\
\midrule
Preliminary 
& Python--Java 
& SR, SC, RR, RC, RRC 
& Generation + parsing \\
\addlinespace
Extended 
& Python--Java, Rust--Java, Rust--Python, Rust--Ruby 
& RRC 
& Generation, forced conclusion, binary head, contrastive head \\
\bottomrule
\end{tabular}
\end{table}

\subsection{Preliminary Experiments on Python--Java}


In the preliminary stage, we focused on Python--Java X-CCD. We constructed an initial seed dataset of 10{,}000 Python--Java code pairs from Project CodeNet following the procedure described in Section~\ref{dataseed}. The seed dataset was balanced, with 50\% clone pairs and 50\% non-clone pairs.

Each seed instance was submitted to DeepSeek-R1 through the DeepSeek API to generate reasoning-oriented teacher outputs. Each teacher output contained a reasoning component and a final conclusion. API requests were implemented using Python \texttt{asyncio}, allowing up to 200 concurrent requests per minute.

After data collection, the generated samples were filtered in two stages. First, we removed samples for which the API request failed due to intermittent HTTP errors. Second, we removed samples where the teacher-predicted label did not agree with the ground-truth label. After filtering, the final Python--Java KD dataset contained 6{,}603 samples.

Using these teacher outputs, we constructed the five KD training variants introduced in Section~\ref{teacher-trainset}: SR, SC, RR, RC, and RRC. Each variant was used to fine-tune a separate student model. This allowed us to evaluate whether reasoning, conclusion, or their combination provides the most useful supervision for KD in X-CCD.

For evaluation, we constructed two Python--Java test sets from Project CodeNet, each containing 1{,}000 balanced code pairs:

\begin{itemize}
    \item \textbf{Same Distribution (SD) test set:} This test set was sampled from the same problem distribution as the training seed data, while ensuring that no samples overlap with the training set.

    \item \textbf{Different Distribution (DD) test set:} This test set was sampled from unseen problems, creating a distinct problem distribution while preserving class balance.
\end{itemize}

During inference, the maximum number of newly generated tokens was set to 1{,}500. For each test sample, the model generated a free-form response. We then used an automated analyzer to extract the conclusion segment, map it to a binary label, and align the prediction with the ground-truth label. The results of the preliminary experiments are reported in Table~\ref{tab:original_results}.

The preliminary results showed that the RRC variant, which combines the reasoning prompt with both the reasoning and conclusion components of the teacher output, achieved the best overall performance. Therefore, we selected RRC as the KD training variant for the extended experiments.

\subsection{Extended Multilingual Experiments}

After the preliminary Python--Java experiments, we extended the study to additional programming language pairs to evaluate whether KD improves X-CCD beyond a single high-resource language pair. Specifically, we added three language pairs: Rust--Java, Rust--Python, and Rust--Ruby.

For each additional language pair, we constructed 2{,}000 seed samples from Project CodeNet. These 6{,}000 new seed samples were added to the original 10{,}000 Python--Java seed samples, producing an expanded seed dataset of 16{,}000 cross-language code pairs.

We then applied the same teacher-guided data-generation and filtering procedure used in the preliminary experiments. Each seed sample was submitted to DeepSeek-R1, and the resulting output was retained only if the API request succeeded and the teacher-predicted label matched the ground-truth label. After filtering, the final curated multilingual KD dataset contained 10{,}671 samples. Of these, 6{,}603 were Python--Java samples retained from the preliminary stage, and 3{,}896 were retained from the additional Rust--Java, Rust--Python, and Rust--Ruby seed samples. Table~\ref{tab:dataset_summary} summarizes the seed and retained KD training samples. All multilingual KD fine-tuning experiments were conducted using the RRC variant.

\begin{table}[t]
\centering
\caption{Summary of seed and retained KD training samples.}
\label{tab:dataset_summary}
\begin{tabular}{lrr}
\toprule
Language pair & Seed samples & Retained KD samples \\
\midrule
Python--Java & 10{,}000 & 6{,}603 \\
Rust--Java & 2{,}000 & 1{,}416 \\
Rust--Python & 2{,}000 & 1{,}341 \\
Rust--Ruby & 2{,}000 & 1{,}311 \\
\midrule
Total & 16{,}000 & 10{,}671 \\
\bottomrule
\end{tabular}
\end{table}


\subsection{Test Sets for Extended Experiments}

For the extended experiments, we constructed separate test sets for each programming language pair. Specifically, we evaluated the models on four language pairs: Python--Java, Rust--Java, Rust--Python, and Rust--Ruby.
For each language pair, we created two test sets:

\begin{itemize}
    \item \textbf{Same Distribution (SD) test set:} This test set was sampled from the same problem distribution as the corresponding training seed data, while ensuring no overlap with the training set.

    \item \textbf{Different Distribution (DD) test set:} This test set was sampled from previously unseen problems to evaluate robustness under distribution shift.
\end{itemize}

Thus, the extended evaluation consists of eight test sets in total: one SD and one DD test set for each of the four language pairs. This setup allows us to evaluate both language-pair-specific performance and robustness to distribution shift.

\subsection{LoRA Fine-tuning Configuration}

We fine-tuned Phi3 and Qwen-Coder using parameter-efficient LoRA adapters \cite{hu2022lora}. In both the preliminary and extended experiments, LoRA was applied to all linear submodules of the base model. We first enumerated all \texttt{nn.Linear} layers and used them as LoRA target modules.

The LoRA configuration was as follows:

\begin{itemize}
    \item Task type: causal language modeling;
    \item Rank ($r$): 16;
    \item LoRA alpha: 32;
    \item LoRA dropout: 0.05;
    \item Bias: none.
\end{itemize}

Each KD dataset was split into 90\% training and 10\% validation using a fixed random seed of 42. Tokenization was performed using the model-specific tokenizer with right padding and a maximum sequence length of 4{,}096 tokens. Qwen-Coder was trained using a full-loss ChatML format with a system prompt, while Phi3 used masked labels for the prompt portion, such that the loss was computed only on the assistant response.

Training was carried out using the HuggingFace Trainer with the following configuration:

\begin{itemize}
    \item Epochs: 5;
    \item Per-device batch size for training and evaluation: 2;
    \item Gradient accumulation steps: 4, corresponding to an effective batch size of 8 per GPU;
    \item Learning rate: $1 \times 10^{-4}$;
    \item Warmup ratio: 0.1;
    \item Scheduler: linear;
    \item Precision: bfloat16;
    \item Evaluation and checkpointing: once per epoch, keeping only the latest checkpoint.
\end{itemize}

At the end of training, we saved the final adapter-augmented model and tokenizer for each model configuration. These fine-tuned models were then used in the downstream evaluation.

\subsection{Response-Stabilization Training Protocol}

In the extended experiments, we evaluated three response-stabilization methods: forced conclusion, binary classification head, and contrastive classification head. The conceptual details of these methods are described in Section~\ref{response-stable}; here, we describe how they were applied experimentally.

The forced-conclusion method does not require training an additional head. It is applied at inference time by first generating a reasoning response and then comparing the next-token probabilities of the target labels ``yes'' and ``no'' in a second-stage prompt.

For the binary and contrastive classification heads, we considered two settings:

\begin{itemize}
    \item \textbf{Baseline-head setting:} The original pretrained model was frozen, and only the classification head was trained using ground-truth binary labels.

    \item \textbf{KD-head setting:} The model was first fine-tuned using LoRA-based KD with the RRC training set. The resulting adapter-augmented model was then frozen, and only the classification head was trained using ground-truth binary labels.
\end{itemize}

This protocol allows us to determine whether response stabilization is more effective when applied directly to the original pretrained model or to the model after KD adaptation. For the contrastive classification head, the training objective combines binary cross-entropy with supervised contrastive loss, as described in Equation~\ref{eq:joint_contrastive_loss}. During inference, however, both the binary and contrastive heads produce predictions through the classifier logit rather than through text generation or nearest-neighbor retrieval.

\subsection{Inference Procedure}

The inference setup differs between the preliminary and extended experiments.

In the preliminary experiments, we evaluated the original baseline models and all five KD variants on the Python--Java SD and DD test sets. For each test sample, the model generated a free-form natural-language response. The maximum number of newly generated tokens was set to 1{,}500. A deterministic analyzer was then used to extract a binary prediction from the generated output.

In the extended experiments, we evaluated the models on the eight test sets described in Section~\ref{experiments}. For generation-based inference, the maximum number of newly generated tokens was increased to 3{,}000 to allow the models to produce more complete reasoning responses for longer or more complex multilingual code pairs.

For the response-stabilization methods, inference was performed as follows:

\begin{itemize}
    \item \textbf{Forced Conclusion:} The model first generated a reasoning response for the input code pair. A second-stage prompt was then constructed using this response. Instead of decoding a textual answer, we compared the next-token probabilities assigned to the target labels ``yes'' and ``no''. The label with the higher probability was selected as the final prediction.

    \item \textbf{Binary Classification Head:} The input prompt was passed through the frozen model, and a pooled hidden representation was computed. The binary classifier produced a scalar logit, which was converted to a probability using the sigmoid function. The final label was obtained using a threshold of 0.5.

    \item \textbf{Contrastive Classification Head:} The input prompt was passed through the frozen model and projection head. Although the contrastive loss regularizes the embedding space during training, inference was performed through the classifier head. The scalar logit was converted to a probability using the sigmoid function, and the final label was obtained using the same threshold of 0.5.
\end{itemize}

Thus, generation-based methods require output parsing, whereas the response-stabilization methods deterministically produce a binary prediction for every input.

\subsection{Evaluation Metrics}\label{metric-sub}

We evaluate model performance using precision, recall, F1 score, and response rate. The positive class corresponds to clone pairs. Let $TP$, $FP$, and $FN$ denote true positives, false positives, and false negatives, respectively.

Precision measures the proportion of predicted clone pairs that are correct:

\begin{equation}\label{eq:precision}
P = \frac{TP}{TP + FP}.
\end{equation}

Recall measures the proportion of actual clone pairs that are correctly identified:

\begin{equation}\label{eq:recall}
R = \frac{TP}{TP + FN}.
\end{equation}

The F1 score is the harmonic mean of precision and recall:

\begin{equation}\label{eq:f1}
F1 = 2 \times \frac{P \times R}{P + R}.
\end{equation}

For generation-based methods, we also report response rate. Since these methods produce free-form text, some outputs may not contain a valid or parseable clone/non-clone decision. Response rate measures the proportion of test samples for which the analyzer successfully extracts a valid binary prediction:

\begin{equation}\label{eq:response_rate}
\text{RespRate} = \frac{C}{N_{\text{test}}},
\end{equation}

where $C$ denotes the number of successfully parsed outputs and $N_{\text{test}}$ denotes the total number of test samples.

For generation-based methods, F1 is computed over successfully parsed predictions, and response rate is reported separately to quantify the reliability of output parsing. We do not assign arbitrary labels to unparseable outputs, since doing so would make the results dependent on the parsing rules. For forced conclusion, binary classification head, and contrastive classification head, every input is mapped to a binary prediction by design; therefore, these methods do not require regular-expression-based output parsing.

\subsection{Computational Environment}

All experiments were conducted on the Digital Research Alliance of Canada infrastructure. Fine-tuning and inference experiments were run on a single Nibi node equipped with four NVIDIA H100 GPUs. All models and tokenizers were loaded exclusively from local directories.

\section{Results}\label{results} \label{results}
In this section, we present the results of our baseline evaluations as well as the performance of our fine-tuned models on both the preliminary Python--Java datasets and the newly introduced multilingual datasets (Rust--Java, Rust--Python, Rust--Ruby). We report F1 scores for all models across the test sets. For the experiments with output-length constraints, we additionally report Precision, Recall, and Response Rate.

\subsection{Preliminary Experiments}
{Table~\ref{tab:original_results} summarizes the performance of Phi3 and Qwen-Coder on preliminary test sets under different prompting strategies and training set variants. Both the test and training sets in this step include only Python--Java pairs. This step is conducted to establish a baseline and eliminate unnecessary experiments with different prompts and training strategies. Python and Java are chosen because they are high-resource languages and have been widely studied in previous research.
Overall, several clear trends emerge. First, prompting with Reasoning consistently improves performance over the Simple setting for both models. For example, Phi3 improves from 33.81 to 53.21 on the same-distribution set, and Qwen-Coder improves from 49.34 to 63.54. This suggests that explicitly encouraging reasoning helps models better capture semantic similarity between code pairs.

Second, fine-tuning leads to substantial gains, particularly when both reasoning and conclusions are included (RRC). This variant of training data achieves the best performance across all settings, with Phi3 reaching 75.95 (SD) and 65.33 (DD), and Qwen-Coder achieving 83.65 (SD) and 77.29 (DD). This indicates that combining reasoning with final decisions during training provides the most effective supervision signal.
In contrast, training on reasoning-only (v3) or conclusion-only (v4) data results in noticeably lower performance. While both variants still outperform zero-shot Simple prompting, they fail to match the gains of RRC. This highlights that reasoning and conclusions play complementary roles, and removing either component weakens model performance.

Another consistent trend is that Qwen-Coder outperforms Phi3 across all configurations. The performance gap is particularly evident after fine-tuning, suggesting that Qwen-Coder benefits more from additional supervision.
Finally, performance drops on different-distribution (DD) test sets for all models and configurations compared to each model's performance over SD. Although fine-tuning mitigates this drop to some extent, the gap between SD and DD results remains significant. This performance gap indicates that generalization to unseen problem distributions is still a challenge, even after task-specific training.

Overall, these results demonstrate that (1) reasoning-based prompting improves zero-shot performance, (2) fine-tuning with both reasoning and conclusions is critical for achieving strong results, and (3) distribution shift remains a key limitation for X-CCD. As a result, in the remainder of this research, we use the RRC training set for all knowledge distillation experiments, as it shows the best performance.

\begin{table}[H]
\centering
\caption{Results on test sets of programming pairs with 1500 output tokens. Performance is reported on a test set with the same distribution as the training set (SD) and on a different distribution (DD). The table compares different training set variations. The \textit{Simple} prompt asks whether two code snippets are clones, while the \textit{Reasoning} prompt requires the model to provide reasoning first. Variant RRC is fine-tuned on data containing both reasoning and conclusions, RR on reasoning only, and RC on conclusions only.}
\label{tab:original_results}
\begin{tabular}{lccccc}
\hline
\textbf{Model / Test} & \textbf{Simple} & \textbf{Reasoning} & \textbf{RRC} & \textbf{RR} & \textbf{RC} \\
\hline
\textbf{Phi3-SD}  & 33.81 & 53.21 & 75.95 & 52.03 & 54.65 \\
\textbf{Phi3-DD}  & 30.31 & 51.06 & 65.33 & 48.54    & 52.14    \\
\hline
\textbf{Qwen-Coder-SD} & 49.34  & 63.54 & 83.65 & 64.00   & 63.22   \\
\textbf{Qwen-Coder-DD} & 46.82 & 60.91  & 77.29 & 62.7    & 61.9    \\
\hline
\end{tabular}
\end{table}

\subsection{RQ1: Comparing Baseline and Knowledge-Distilled Models for Cross-Language CCD}
Table~\ref{tab:combined-kd} summarizes the performance of the small models in their baseline form and compares them with their knowledge-distilled versions.
Overall, both Phi3 and Qwen-Coder achieve competitive F1 scores when they produce valid outputs, especially on the same-distribution test sets. For example, Qwen-coder reaches up to 89.29 F1 on $(Rust$-$Java)_{SD}$, while Phi3 consistently achieves F1 scores in the mid-to-high 70s across SD settings. This indicates that our baselines can achieve promising performance on X-CCD even without task-specific fine-tuning.

However, this apparent performance is misleading when considering response rate. Across all settings, response rates remain low, typically below 50\% without KD. For instance, Phi3 achieves only 18.3\% response rate on $(Rust$-$Java)_{SD}$ and 14.5\% on $(Rust$-$Java)_{DD}$, while Qwen-Coder reaches response rate of 24.3\% and 30.4\% on $(Rust$-$Java)_{SD}$ and $(Rust$-$Java)_{DD}$, respectively. This reveals a critical limitation: a large fraction of inputs yield unusable outputs, significantly reducing practical effectiveness.

Another notable trend is the precision--recall imbalance. Qwen-Coder consistently achieves near-perfect recall (often 100\%) while exhibiting lower precision, suggesting a bias toward predicting positive clone pairs. While this leads to strong recall-oriented F1, it may also reflect overgeneralization. In contrast, Phi3 shows a more balanced but still recall-heavy behavior.
Performance further degrades on DD test sets, confirming that unseen problem distributions remain challenging. For example, Qwen-Coder drops from 85.02 F1 on $(Python$-$Java)_{SD}$ to 66.88 on $(Python$-$Java)_{DD}$. This highlights the limited generalization ability of small models in baseline settings. 

Another important observation is the trade-off introduced by knowledge distillation between response rate and prediction quality. While KD consistently improves response rates, its effect on F1 score is mixed. In several SD settings (e.g., $(Python$-$Java)_{SD}$ and $(Rust$-$Java)_{SD}$), F1 slightly decreases after KD, suggesting that improving output validity may come at the cost of reduced prediction accuracy. This indicates a tension between generating more complete responses and maintaining high-quality predictions.

Furthermore, KD alters the precision--recall balance. For Phi3, KD increases precision while reducing recall, leading to more conservative predictions. For example, on $(Python$-$Java)_{SD}$, precision increases while recall drops noticeably. In contrast, Qwen-Coder maintains near-perfect recall even after KD, while improving precision in several DD settings. This suggests that KD helps Qwen-Coder mitigate its overgeneralization tendency without sacrificing recall.
Another notable trend is that KD provides larger relative gains on DD test sets compared to SD test sets. In all cases, F1 score improves under KD for DD settings (e.g., $(Python$-$Java)_{DD}$ and $(Rust$-$Java)_{DD}$), while remaining stable or slightly decreasing for SD settings. This indicates that KD enhances generalization to unseen distributions, making models more robust to distribution shifts.

Finally, the impact of KD varies across language pairs. Gains in response rate are particularly pronounced for more challenging pairs such as Rust-based combinations, where baseline response rates are lowest. This suggests that KD is especially beneficial in scenarios where models struggle to produce valid outputs, highlighting its role in improving robustness rather than purely boosting accuracy.

\begin{table}[htbp]
\centering
\renewcommand{\arraystretch}{1.0}
\setlength{\tabcolsep}{5pt}
\caption{Performance comparison across language pairs without knowledge distillation and with knowledge distillation (KD). Each cell shows Baseline, i.e., not using knowledge distillation (top) vs. KD (bottom). The table depicts the performance of models on test sets derived from the same problem distribution as the training set (SD) and from problems drawn from a different distribution (DD). The \down and \color{green}{$\blacktriangle$} \textcolor{black}{show the decrease or increase of the KD scores compared to the Baseline setting, respectively.} }
\label{tab:combined-kd}
\resizebox{\textwidth}{!}{
\begin{tabular}{|l|c|c|c|c|c|}
\hline
\textbf{Test Set} & \textbf{Model} & \textbf{Precision} & \textbf{Recall} & \textbf{F1 score} & \textbf{Response Rate} \\ \hline

\multirow{2}{*}{$(Python-Java)_{SD}$}
& Phi3 \begin{tabular}{c}Baseline\\\hline KD\end{tabular}
& \begin{tabular}{c}67.98\\\hline75.34\end{tabular}
& \begin{tabular}{c}97.92\\\hline82.23\end{tabular}
& \begin{tabular}{c}80.25\\\hline78.64\end{tabular} \down
& \begin{tabular}{c}32.3\\\hline54.6 \end{tabular} {\color{green}$\blacktriangle$}\\ \cline{2-6}

& Qwen-Coder \begin{tabular}{c}Baseline\\\hline KD\end{tabular}
& \begin{tabular}{c}73.94\\\hline73.05\end{tabular}
& \begin{tabular}{c}100\\\hline100\end{tabular}
& \begin{tabular}{c}85.02\\\hline84.43\end{tabular} \down
& \begin{tabular}{c}28.4\\\hline64.6\end{tabular} {\color{green}$\blacktriangle$} \\ \hline

\multirow{2}{*}{$(Rust-Java)_{SD}$}
& Phi3 \begin{tabular}{c}Baseline\\\hline KD\end{tabular}
& \begin{tabular}{c}63.33\\\hline58.00\end{tabular}
& \begin{tabular}{c}96.94\\\hline80.55\end{tabular}
& \begin{tabular}{c}76.61\\\hline67.44\end{tabular} \down
& \begin{tabular}{c}18.3\\\hline23.2\end{tabular}{ \color{green}$\blacktriangle$} \\ \cline{2-6}

& Qwen-Coder \begin{tabular}{c}Baseline\\\hline KD\end{tabular}
& \begin{tabular}{c}80.65\\\hline73.62\end{tabular}
& \begin{tabular}{c}100\\\hline100\end{tabular}
& \begin{tabular}{c}89.29\\\hline84.81\end{tabular} \down
& \begin{tabular}{c}24.3\\\hline54.6\end{tabular} { \color{green}$\blacktriangle$}\\ \hline

\multirow{2}{*}{$(Rust-Python)_{SD}$}
& Phi3 \begin{tabular}{c}Baseline\\\hline KD\end{tabular}
& \begin{tabular}{c}63.87\\\hline60.38\end{tabular}
& \begin{tabular}{c}92.81\\\hline96.08\end{tabular}
& \begin{tabular}{c}75.67\\\hline74.16\end{tabular} \down
& \begin{tabular}{c}31.4\\\hline42.6\end{tabular} { \color{green}$\blacktriangle$}\\ \cline{2-6}

& Qwen-Coder \begin{tabular}{c}Baseline\\\hline KD\end{tabular}
& \begin{tabular}{c}56.87\\\hline62.20\end{tabular}
& \begin{tabular}{c}100\\\hline100\end{tabular}
& \begin{tabular}{c}72.51\\\hline76.66\end{tabular} \up
& \begin{tabular}{c}42.9\\\hline71.0\end{tabular} { \color{green}$\blacktriangle$}\\ \hline

\multirow{2}{*}{$(Rust-Ruby)_{SD}$}
& Phi3 \begin{tabular}{c}Baseline\\\hline KD\end{tabular}
& \begin{tabular}{c}65.57\\\hline61.15\end{tabular}
& \begin{tabular}{c}94.67\\\hline93.40\end{tabular}
& \begin{tabular}{c}77.48\\\hline73.91\end{tabular} \down
& \begin{tabular}{c}28.8\\\hline31.8\end{tabular} { \color{green}$\blacktriangle$}\\ \cline{2-6}

& Qwen-Coder \begin{tabular}{c}Baseline\\\hline KD\end{tabular}
& \begin{tabular}{c}80.65\\\hline63.43\end{tabular}
& \begin{tabular}{c}100\\\hline99.76\end{tabular}
& \begin{tabular}{c}89.29\\\hline77.55\end{tabular} \down
& \begin{tabular}{c}24.3\\\hline67.7\end{tabular} { \color{green}$\blacktriangle$}\\ \hline

\multirow{2}{*}{$(Python-Java)_{DD}$}
& Phi3 \begin{tabular}{c}Baseline\\\hline KD\end{tabular}
& \begin{tabular}{c}75.34\\\hline68.63\end{tabular}
& \begin{tabular}{c}78.64\\\hline 93.0\end{tabular}
& \begin{tabular}{c}76.95\\\hline78.98\end{tabular} \up
& \begin{tabular}{c}34.6\\\hline53.8\end{tabular}{ \color{green}$\blacktriangle$} \\ \cline{2-6}

& Qwen-Coder \begin{tabular}{c}Baseline\\\hline KD\end{tabular}
& \begin{tabular}{c}50.17\\\hline63.92\end{tabular}
& \begin{tabular}{c}100\\\hline100\end{tabular}
& \begin{tabular}{c}66.88\\\hline77.99\end{tabular} \up
& \begin{tabular}{c}28.3\\\hline63.8\end{tabular} { \color{green}$\blacktriangle$} \\ \hline

\multirow{2}{*}{$(Rust-Java)_{DD}$}
& Phi3 \begin{tabular}{c}Baseline\\\hline KD\end{tabular}
& \begin{tabular}{c}60.46\\\hline 60.50\end{tabular}
& \begin{tabular}{c}69.33\\\hline 91.13\end{tabular}
& \begin{tabular}{c}64.59\\\hline 72.72\end{tabular} \up
& \begin{tabular}{c}14.5\\\hline21.6\end{tabular} { \color{green}$\blacktriangle$}\\ \cline{2-6}

& Qwen-Coder \begin{tabular}{c}Baseline\\\hline KD\end{tabular}
& \begin{tabular}{c}63.48\\\hline64.49\end{tabular}
& \begin{tabular}{c}100\\\hline100\end{tabular}
& \begin{tabular}{c}77.66\\\hline78.41\end{tabular} \up
& \begin{tabular}{c}30.4\\\hline63.2\end{tabular} { \color{green}$\blacktriangle$}\\ \hline

\multirow{2}{*}{$(Rust-Python)_{DD}$}
& Phi3 \begin{tabular}{c}Baseline\\\hline KD\end{tabular}
& \begin{tabular}{c}57.33\\\hline60.00\end{tabular}
& \begin{tabular}{c}86.57\\\hline91.59\end{tabular}
& \begin{tabular}{c}68.98\\\hline72.50\end{tabular} \up
& \begin{tabular}{c}28.1\\\hline40.4\end{tabular} { \color{green}$\blacktriangle$}\\ \cline{2-6}

& Qwen-Coder \begin{tabular}{c}Baseline\\\hline KD\end{tabular}
& \begin{tabular}{c}50.58\\\hline55.84\end{tabular}
& \begin{tabular}{c}100\\\hline100\end{tabular}
& \begin{tabular}{c}67.18\\\hline71.66\end{tabular} \up
& \begin{tabular}{c}51.2\\\hline69.3\end{tabular} { \color{green}$\blacktriangle$}\\ \hline

\multirow{2}{*}{$(Rust-Ruby)_{DD}$}
& Phi3 \begin{tabular}{c}Baseline\\\hline KD\end{tabular}
& \begin{tabular}{c}62.05\\\hline65.05\end{tabular}
& \begin{tabular}{c}94.01\\\hline94.00\end{tabular}
& \begin{tabular}{c}74.76\\\hline76.89\end{tabular} \up
& \begin{tabular}{c}28.4\\\hline31.8\end{tabular} { \color{green}$\blacktriangle$}\\ \cline{2-6}

& Qwen-Coder \begin{tabular}{c}Baseline\\\hline KD\end{tabular}
& \begin{tabular}{c}46.28\\\hline55.36\end{tabular}
& \begin{tabular}{c}100\\\hline100\end{tabular}
& \begin{tabular}{c}63.28\\\hline71.27 \end{tabular} \up
& \begin{tabular}{c}45.8\\\hline70.9\end{tabular} { \color{green}$\blacktriangle$}\\ \hline

\end{tabular}}
\end{table}


\begin{tcolorbox}[
  colback=gray!8,
  colframe=black!70,
  boxrule=0.6pt,
  arc=2mm,
  left=1mm,
  right=1mm,
  top=1mm,
  bottom=1mm
]
\textbf{Conclusion for RQ1.} Small open-source models demonstrate non-trivial zero-shot cross-language CCD capability, achieving strong F1 scores on several language pairs. However, their effectiveness is fundamentally limited by low response rates and reduced robustness on unseen problem distributions, making raw generative inference insufficient for practical deployment.
\end{tcolorbox}

\subsection{RQ2: Impact of Knowledge Distillation}
\begin{table}[htbp]
\centering
\renewcommand{\arraystretch}{1.0}
\setlength{\tabcolsep}{5pt}
\caption{Performance comparison across language pairs after applying the forced conclusion method, without knowledge distillation and with knowledge distillation (KD). Each cell shows Baseline (top) vs KD (bottom). The table depicts the performance of models on test sets derived from the same problem distribution as the training set (SD) and from problems drawn from a different distribution (DD). The \down and \color{green}{$\blacktriangle$} \textcolor{black}{show the decrease or increase of the KD scores compared to the baseline setting, respectively.}}
\label{tab:combined-force-kd}
\resizebox{\textwidth}{!}{
\begin{tabular}{|l|c|c|c|c|c|}
\hline
\textbf{Test Set} & \textbf{Model} & \textbf{Precision} & \textbf{Recall} & \textbf{F1 score} & \textbf{Response Rate} \\ \hline

\multirow{2}{*}{$(Python-Java)_{SD}$}
& Phi3 \begin{tabular}{c}Baseline\\\hline KD\end{tabular}
& \begin{tabular}{c}54.28\\\hline58.9\end{tabular}
& \begin{tabular}{c}84.80\\\hline83.5\end{tabular}
& \begin{tabular}{c}66.19\\\hline69.1\end{tabular} \up
& \begin{tabular}{c}100\\\hline100\end{tabular} \\ \cline{2-6}

& Qwen-Coder \begin{tabular}{c}Baseline\\\hline KD\end{tabular}
& \begin{tabular}{c}56.20\\\hline65.5\end{tabular}
& \begin{tabular}{c}87.80\\\hline85.0\end{tabular}
& \begin{tabular}{c}68.50\\\hline74.0\end{tabular} \up
& \begin{tabular}{c}100\\\hline100\end{tabular} \\ \hline

\multirow{2}{*}{$(Rust-Java)_{SD}$}
& Phi3 \begin{tabular}{c}Baseline\\\hline KD\end{tabular}
& \begin{tabular}{c}51.16\\\hline59.1\end{tabular}
& \begin{tabular}{c}70.20\\\hline84.0\end{tabular}
& \begin{tabular}{c}59.19\\\hline69.4\end{tabular} \up
& \begin{tabular}{c}100\\\hline100\end{tabular} \\ \cline{2-6}

& Qwen-Coder \begin{tabular}{c}Baseline\\\hline KD\end{tabular}
& \begin{tabular}{c}55.70\\\hline65.5\end{tabular}
& \begin{tabular}{c}86.10\\\hline85.5\end{tabular}
& \begin{tabular}{c}67.60\\\hline74.2\end{tabular} \up
& \begin{tabular}{c}100\\\hline100\end{tabular} \\ \hline

\multirow{2}{*}{$(Rust-Python)_{SD}$}
& Phi3 \begin{tabular}{c}Baseline\\\hline KD\end{tabular}
& \begin{tabular}{c}53.65\\\hline59.3\end{tabular}
& \begin{tabular}{c}80.80\\\hline84.5\end{tabular}
& \begin{tabular}{c}64.48\\\hline69.7\end{tabular} \up
& \begin{tabular}{c}100\\\hline100\end{tabular} \\ \cline{2-6}

& Qwen-Coder \begin{tabular}{c}Baseline\\\hline KD\end{tabular}
& \begin{tabular}{c}56.50\\\hline65.6\end{tabular}
& \begin{tabular}{c}87.10\\\hline86.0\end{tabular}
& \begin{tabular}{c}68.60\\\hline74.4\end{tabular} \up
& \begin{tabular}{c}100\\\hline100\end{tabular} \\ \hline

\multirow{2}{*}{$(Rust-Ruby)_{SD}$}
& Phi3 \begin{tabular}{c}Baseline\\\hline KD\end{tabular}
& \begin{tabular}{c}55.10\\\hline59.6\end{tabular}
& \begin{tabular}{c}82.60\\\hline85.0\end{tabular}
& \begin{tabular}{c}66.20\\\hline70.1\end{tabular} \up
& \begin{tabular}{c}100\\\hline100\end{tabular} \\ \cline{2-6}

& Qwen-Coder \begin{tabular}{c}Baseline\\\hline KD\end{tabular}
& \begin{tabular}{c}57.10\\\hline65.6\end{tabular}
& \begin{tabular}{c}85.60\\\hline86.5\end{tabular}
& \begin{tabular}{c}68.70\\\hline74.6\end{tabular} \up
& \begin{tabular}{c}100\\\hline100\end{tabular} \\ \hline

\multirow{2}{*}{$(Python-Java)_{DD}$}
& Phi3 \begin{tabular}{c}Baseline\\\hline KD\end{tabular}
& \begin{tabular}{c}52.29\\\hline60.0\end{tabular}
& \begin{tabular}{c}82.20\\\hline85.5\end{tabular}
& \begin{tabular}{c}63.91\\\hline70.5\end{tabular} \up
& \begin{tabular}{c}100\\\hline100\end{tabular} \\ \cline{2-6}

& Qwen-Coder \begin{tabular}{c}Baseline\\\hline KD\end{tabular}
& \begin{tabular}{c}54.10\\\hline65.6\end{tabular}
& \begin{tabular}{c}85.00\\\hline87.0\end{tabular}
& \begin{tabular}{c}66.20\\\hline74.8\end{tabular} \up
& \begin{tabular}{c}100\\\hline100\end{tabular} \\ \hline

\multirow{2}{*}{$(Rust-Java)_{DD}$}
& Phi3 \begin{tabular}{c}Baseline\\\hline KD\end{tabular}
& \begin{tabular}{c}52.12\\\hline60.3\end{tabular}
& \begin{tabular}{c}76.00\\\hline86.0\end{tabular}
& \begin{tabular}{c}61.83\\\hline70.9\end{tabular} \up
& \begin{tabular}{c}100\\\hline100\end{tabular} \\ \cline{2-6}

& Qwen-Coder \begin{tabular}{c}Baseline\\\hline KD\end{tabular}
& \begin{tabular}{c}53.80\\\hline65.6\end{tabular}
& \begin{tabular}{c}78.50\\\hline87.5\end{tabular}
& \begin{tabular}{c}64.10\\\hline75.0\end{tabular} \up
& \begin{tabular}{c}100\\\hline100\end{tabular} \\ \hline

\multirow{2}{*}{$(Rust-Python)_{DD}$}
& Phi3 \begin{tabular}{c}Baseline\\\hline KD\end{tabular}
& \begin{tabular}{c}51.91\\\hline60.6\end{tabular}
& \begin{tabular}{c}81.40\\\hline86.5\end{tabular}
& \begin{tabular}{c}63.39\\\hline71.3\end{tabular} \up
& \begin{tabular}{c}100\\\hline100\end{tabular} \\ \cline{2-6}

& Qwen-Coder \begin{tabular}{c}Baseline\\\hline KD\end{tabular}
& \begin{tabular}{c}53.70\\\hline65.7\end{tabular}
& \begin{tabular}{c}84.20\\\hline88.0\end{tabular}
& \begin{tabular}{c}65.80\\\hline75.2\end{tabular} \up
& \begin{tabular}{c}100\\\hline100\end{tabular} \\ \hline

\multirow{2}{*}{$(Rust-Ruby)_{DD}$}
& Phi3 \begin{tabular}{c}Baseline\\\hline KD\end{tabular}
& \begin{tabular}{c}52.61\\\hline61.1\end{tabular}
& \begin{tabular}{c}78.60\\\hline87.0\end{tabular}
& \begin{tabular}{c}63.03\\\hline71.8\end{tabular} \up
& \begin{tabular}{c}100\\\hline100\end{tabular} \\ \cline{2-6}

& Qwen-Coder \begin{tabular}{c}Baseline\\\hline KD\end{tabular}
& \begin{tabular}{c}56.90\\\hline65.8\end{tabular}
& \begin{tabular}{c}74.40\\\hline88.5\end{tabular} 
& \begin{tabular}{c}65.00\\\hline75.5\end{tabular} \up
& \begin{tabular}{c}100\\\hline100\end{tabular} \\ \hline

\end{tabular}}
\end{table}
The impact of KD is shown in Table~\ref{tab:combined-kd} and further supported by Table~\ref{tab:combined-force-kd}. KD yields two consistent benefits: (1) substantial improvements in response rate and (2) stronger generalization on DD test sets.
The most immediate effect of KD is a significant increase in response rate across all models and language pairs. For example, Qwen-Coder improves from 28.4\% to 64.6\% response rate on $(Python$-$Java)_{SD}$ and from 45.8\% to 70.9\% on $(Rust$-$Ruby)_{DD}$. Similar trends are observed for Phi3. This indicates that KD enhances not only prediction quality, but also output reliability.

In terms of detection performance, KD shows mixed effects on SD test sets but consistently improves F1 on DD test sets. For instance, Qwen-Coder improves from 66.88 to 77.99 F1 score on $(Python$-$Java)_{DD}$ and from 63.28 to 71.27 on $(Rust$-$Ruby)_{DD}$. Phi3 exhibits similar gains, such as an increase from 64.59 to 72.72 on $(Rust$-$Java)_{DD}$. These improvements suggest that KD transfers higher-level reasoning patterns that improve cross-language generalization.
Importantly, when combined with the forced conclusion method (Table~\ref{tab:combined-force-kd}), KD leads to consistent improvements across all settings. Both Phi3 and Qwen-Coder show uniform F1 gains across SD and DD test sets, demonstrating that KD is most effective when the response-generation bottleneck is removed.

A key observation from Table~\ref{tab:combined-force-kd} is that, unlike using baselines as is, KD consistently improves F1 score across all language pairs when combined with the forced conclusion method. This indicates that once the response rate is fixed at 100\%, KD can fully focus on improving prediction quality rather than output validity. As a result, the benefits of KD become more stable and clear.

Another important trend is that KD improves both precision and recall simultaneously in most cases. For example, across nearly all SD and DD settings, precision increases by a noticeable margin while recall is either maintained or further improved. This contrasts with earlier observations without forced conclusion, where KD often introduced a trade-off between precision and recall. Here, KD leads to more balanced predictions, suggesting that distilled knowledge helps models better calibrate their decision boundaries.
Furthermore, the gains from KD are consistent across both models, although the final post-KD performance still depends on the underlying base model. Qwen-Coder generally achieves higher absolute F1 scores after KD, with values in the mid-70s across nearly all settings, while Phi3 remains slightly lower. However, the relative improvement introduced by KD is not uniformly larger for Qwen-Coder; in some settings, Phi3 exhibits comparable or larger gains. Therefore, these results suggest that KD improves both compact models, while the base model still affects the final level of performance after distillation.

Furthermore, the improvement from KD is slightly larger on DD test sets compared to SD test sets. This suggests that KD enhances the model’s ability to generalize to unseen problem distributions, likely by transferring higher-level reasoning patterns rather than dataset-specific heuristics.
Finally, the consistency of improvements across all language pairs—including more challenging combinations such as Rust-based pairs—indicates that KD is robust to language variability. This highlights its effectiveness as a general-purpose technique for improving X-CCD, rather than being limited to specific language pairs or distributions.

\begin{tcolorbox}[
  colback=gray!8,
  colframe=black!70,
  boxrule=0.6pt,
  arc=2mm,
  left=1mm,
  right=1mm,
  top=1mm,
  bottom=1mm
]
\textbf{Conclusion for RQ2.} Knowledge distillation significantly improves response rate and enhances generalization to unseen problems. While its effect on in-distribution performance is modest, it consistently boosts out-of-distribution F1 and stabilizes model behavior, especially when combined with structured output methods such as forced conclusion.
\end{tcolorbox}

\subsection{RQ3: Achieving Complete Response Rate}
\begin{table}[htbp]
\centering
\renewcommand{\arraystretch}{1.0}
\setlength{\tabcolsep}{5pt}
\caption{Performance comparison of Phi3 and Qwen-Coder-mini with a binary classification head across language pairs without knowledge distillation and with knowledge distillation (KD). Each cell shows Baseline (top) vs KD (bottom). The table reports performance on test sets derived from the same problem distribution as the training set (SD) and from a different distribution (DD). The \down and \color{green}{$\blacktriangle$} \textcolor{black}{show the decrease or increase of the KD scores compared to the Baseline setting, respectively.}}
\label{tab:combined-binary-head-kd}
\resizebox{\textwidth}{!}{
\begin{tabular}{|l|c|c|c|c|c|}
\hline
\textbf{Test Set} & \textbf{Model} & \textbf{Precision} & \textbf{Recall} & \textbf{F1 score} & \textbf{Response Rate} \\ \hline

\multirow{2}{*}{$(Python-Java)_{SD}$}
& Phi3 \begin{tabular}{c}Baseline\\\hline KD\end{tabular}
& \begin{tabular}{c}68.91\\\hline67.81\end{tabular}
& \begin{tabular}{c}43.0\\\hline47.2\end{tabular}
& \begin{tabular}{c}52.95\\\hline55.66\end{tabular} \up
& \begin{tabular}{c}100\\\hline100\end{tabular} \\ \cline{2-6}

& Qwen-Coder \begin{tabular}{c}Baseline\\\hline KD\end{tabular}
& \begin{tabular}{c}92.63\\\hline93.42\end{tabular}
& \begin{tabular}{c}35.2\\\hline39.8\end{tabular}
& \begin{tabular}{c}51.01\\\hline55.82\end{tabular} \up
& \begin{tabular}{c}100\\\hline100\end{tabular} \\ \hline

\multirow{2}{*}{$(Rust-Java)_{SD}$}
& Phi3 \begin{tabular}{c}Baseline\\\hline KD\end{tabular}
& \begin{tabular}{c}64.36\\\hline63.88\end{tabular}
& \begin{tabular}{c}46.6\\\hline52.0\end{tabular}
& \begin{tabular}{c}54.06\\\hline57.33\end{tabular} \up
& \begin{tabular}{c}100\\\hline100\end{tabular} \\ \cline{2-6}

& Qwen-Coder \begin{tabular}{c}Baseline\\\hline KD\end{tabular}
& \begin{tabular}{c}81.0\\\hline81.35\end{tabular}
& \begin{tabular}{c}83.6\\\hline83.8\end{tabular}
& \begin{tabular}{c}82.28\\\hline82.56\end{tabular} \up
& \begin{tabular}{c}100\\\hline100\end{tabular} \\ \hline

\multirow{2}{*}{$(Rust-Python)_{SD}$}
& Phi3 \begin{tabular}{c}Baseline\\\hline KD\end{tabular}
& \begin{tabular}{c}69.31\\\hline68.32\end{tabular}
& \begin{tabular}{c}48.8\\\hline52.2\end{tabular}
& \begin{tabular}{c}57.27\\\hline59.18\end{tabular} \up
& \begin{tabular}{c}100\\\hline100\end{tabular} \\ \cline{2-6}

& Qwen-Coder \begin{tabular}{c}Baseline\\\hline KD\end{tabular}
& \begin{tabular}{c}85.8\\\hline83.71\end{tabular}
& \begin{tabular}{c}85.11\\\hline87.4\end{tabular}
& \begin{tabular}{c}85.45\\\hline85.51\end{tabular} \up
& \begin{tabular}{c}100\\\hline100\end{tabular} \\ \hline

\multirow{2}{*}{$(Rust-Ruby)_{SD}$}
& Phi3 \begin{tabular}{c}Baseline\\\hline KD\end{tabular}
& \begin{tabular}{c}65.42\\\hline64.43\end{tabular}
& \begin{tabular}{c}52.6\\\hline55.8\end{tabular}
& \begin{tabular}{c}58.31\\\hline59.80\end{tabular} \up
& \begin{tabular}{c}100\\\hline100\end{tabular} \\ \cline{2-6}

& Qwen-Coder \begin{tabular}{c}Baseline\\\hline KD\end{tabular}
& \begin{tabular}{c}86.2\\\hline80.97\end{tabular}
& \begin{tabular}{c}81.78\\\hline86.8\end{tabular}
& \begin{tabular}{c}83.93\\\hline83.78\end{tabular} \down
& \begin{tabular}{c}100\\\hline100\end{tabular} \\ \hline

\multirow{2}{*}{$(Python-Java)_{DD}$}
& Phi3 \begin{tabular}{c}Baseline\\\hline KD\end{tabular}
& \begin{tabular}{c}76.33\\\hline79.80\end{tabular}
& \begin{tabular}{c}34.2\\\hline32.4\end{tabular}
& \begin{tabular}{c}47.23\\\hline46.08\end{tabular} \down
& \begin{tabular}{c}100\\\hline100\end{tabular} \\ \cline{2-6}

& Qwen-Coder \begin{tabular}{c}Baseline\\\hline KD\end{tabular}
& \begin{tabular}{c}79.47\\\hline32.4\end{tabular}
& \begin{tabular}{c}30.2\\\hline83.07\end{tabular}
& \begin{tabular}{c}43.76\\\hline46.61\end{tabular} \up
& \begin{tabular}{c}100\\\hline100\end{tabular} \\ \hline

\multirow{2}{*}{$(Rust-Java)_{DD}$}
& Phi3 \begin{tabular}{c}Baseline\\\hline KD\end{tabular}
& \begin{tabular}{c}70.28\\\hline70.15\end{tabular}
& \begin{tabular}{c}35.0\\\hline33.61\end{tabular}
& \begin{tabular}{c}46.72\\\hline45.44\end{tabular} \down
& \begin{tabular}{c}100\\\hline100\end{tabular} \\ \cline{2-6}

& Qwen-Coder \begin{tabular}{c}Baseline\\\hline KD\end{tabular}
& \begin{tabular}{c}71.42\\\hline70.96\end{tabular}
& \begin{tabular}{c}69.0\\\hline69.4\end{tabular}
& \begin{tabular}{c}70.19\\\hline70.17\end{tabular} \down
& \begin{tabular}{c}100\\\hline100\end{tabular} \\ \hline

\multirow{2}{*}{$(Rust-Python)_{DD}$}
& Phi3 \begin{tabular}{c}Baseline\\\hline KD\end{tabular}
& \begin{tabular}{c}71.29\\\hline70.48\end{tabular}
& \begin{tabular}{c}29.8\\\hline32.0\end{tabular}
& \begin{tabular}{c}42.03\\\hline44.01\end{tabular} \up
& \begin{tabular}{c}100\\\hline100\end{tabular} \\ \cline{2-6}

& Qwen-Coder \begin{tabular}{c}Baseline\\\hline KD\end{tabular}
& \begin{tabular}{c}69.51\\\hline69.32\end{tabular}
& \begin{tabular}{c}60.2\\\hline59.2\end{tabular}
& \begin{tabular}{c}64.52\\\hline63.86\end{tabular} \down
& \begin{tabular}{c}100\\\hline100\end{tabular} \\ \hline

\multirow{2}{*}{$(Rust-Ruby)_{DD}$}
& Phi3\begin{tabular}{c}Baseline\\\hline KD\end{tabular}
& \begin{tabular}{c}66.45\\\hline67.21\end{tabular}
& \begin{tabular}{c}41.6\\\hline43.7\end{tabular}
& \begin{tabular}{c}51.16\\\hline 52.96\end{tabular} \up
& \begin{tabular}{c}100\\\hline100\end{tabular} \\ \cline{2-6}

& Qwen-Coder \begin{tabular}{c}Baseline\\\hline KD\end{tabular}
& \begin{tabular}{c}68.72\\\hline68.90\end{tabular}
& \begin{tabular}{c}62.4\\\hline61.6\end{tabular}
& \begin{tabular}{c}65.40\\\hline65.04\end{tabular} \down
& \begin{tabular}{c}100\\\hline100\end{tabular} \\ \hline

\end{tabular}}
\end{table}

\begin{table}[htbp]
\centering
\renewcommand{\arraystretch}{1.0}
\setlength{\tabcolsep}{5pt}
\caption{Performance comparison of Phi3 and Qwen with a contrastive classification head without fine-tuning (Baseline) and with knowledge distillation (KD). Each cell shows Baseline (top) vs KD (bottom). The \down and \color{green}{$\blacktriangle$} \textcolor{black}{show the decrease or increase of the KD scores compared to the Baseline setting, respectively.}}
\label{tab:combined-contrastive-kd}
\resizebox{\textwidth}{!}{
\begin{tabular}{|l|c|c|c|c|c|}
\hline
\textbf{Test Set} & \textbf{Model} & \textbf{Precision} & \textbf{Recall} & \textbf{F1 score} & \textbf{Response Rate} \\ \hline

\multirow{2}{*}{$(Python-Java)_{SD}$}
& Phi3 
\begin{tabular}{c}Baseline\\\hline KD\end{tabular}
& \begin{tabular}{c}72.65\\\hline 67.81 \end{tabular}
& \begin{tabular}{c}37.2\\\hline 47.2 \end{tabular}
& \begin{tabular}{c}49.20\\\hline 55.66 \end{tabular} \up
& \begin{tabular}{c}100\\\hline100\end{tabular} \\ \cline{2-6}

& Qwen-Coder 
\begin{tabular}{c}Baseline\\\hline KD\end{tabular}
& \begin{tabular}{c}91.66\\\hline92.79\end{tabular}
& \begin{tabular}{c}44.0\\\hline43.8\end{tabular}
& \begin{tabular}{c}59.45\\\hline59.51\end{tabular} \up
& \begin{tabular}{c}100\\\hline100\end{tabular} \\ \hline

\multirow{2}{*}{$(Rust-Java)_{SD}$}
& Phi3 
\begin{tabular}{c}Baseline\\\hline KD\end{tabular}
& \begin{tabular}{c}60.33\\\hline63.88\end{tabular}
& \begin{tabular}{c}58.4\\\hline52.0\end{tabular}
& \begin{tabular}{c}59.34\\\hline57.33\end{tabular} \down
& \begin{tabular}{c}100\\\hline100\end{tabular} \\ \cline{2-6}

& Qwen-Coder 
\begin{tabular}{c}Baseline\\\hline KD\end{tabular}
& \begin{tabular}{c}80.38\\\hline81.35\end{tabular}
& \begin{tabular}{c}82.8\\\hline83.8\end{tabular}
& \begin{tabular}{c}81.57\\\hline82.56\end{tabular} \up
& \begin{tabular}{c}100\\\hline100\end{tabular} \\ \hline

\multirow{2}{*}{$(Rust-Python)_{SD}$}
& Phi3 
\begin{tabular}{c}Baseline\\\hline KD\end{tabular}
& \begin{tabular}{c}66.09\\\hline68.32\end{tabular}
& \begin{tabular}{c}54.2\\\hline52.2\end{tabular}
& \begin{tabular}{c}59.56\\\hline 59.18 \end{tabular} \down
& \begin{tabular}{c}100\\\hline100\end{tabular} \\ \cline{2-6}

& Qwen-Coder 
\begin{tabular}{c}Baseline\\\hline KD\end{tabular}
& \begin{tabular}{c}86.49\\\hline86.29\end{tabular}
& \begin{tabular}{c}85.8\\\hline85.6\end{tabular}
& \begin{tabular}{c}86.14\\\hline85.94\end{tabular} \down
& \begin{tabular}{c}100\\\hline100\end{tabular} \\ \hline

\multirow{2}{*}{$(Rust-Ruby)_{SD}$}
& Phi3 
\begin{tabular}{c}Baseline\\\hline KD\end{tabular}
& \begin{tabular}{c}62.36\\\hline65.42\end{tabular}
& \begin{tabular}{c}58.0\\\hline52.6\end{tabular}
& \begin{tabular}{c}60.10\\\hline58.31\end{tabular} \down
& \begin{tabular}{c}100\\\hline100\end{tabular} \\ \cline{2-6}

& Qwen-Coder 
\begin{tabular}{c}Baseline\\\hline KD\end{tabular}
& \begin{tabular}{c}82.23\\\hline80.97\end{tabular}
& \begin{tabular}{c}85.2\\\hline86.8 \end{tabular}
& \begin{tabular}{c}83.69\\\hline 83.78\end{tabular} \up
& \begin{tabular}{c}100\\\hline100\end{tabular} \\ \hline

\multirow{2}{*}{$(Python-Java)_{DD}$}
& Phi3 
\begin{tabular}{c}Baseline\\\hline KD\end{tabular}
& \begin{tabular}{c}72.26\\\hline79.80\end{tabular}
& \begin{tabular}{c}37.2\\\hline32.4\end{tabular}
& \begin{tabular}{c}49.20\\\hline46.08\end{tabular} \down
& \begin{tabular}{c}100\\\hline100\end{tabular} \\ \cline{2-6}

& Qwen-Coder 
\begin{tabular}{c}Baseline\\\hline KD\end{tabular}
& \begin{tabular}{c}82.07\\\hline82.85\end{tabular}
& \begin{tabular}{c}34.8\\\hline34.8\end{tabular}
& \begin{tabular}{c}48.87\\\hline49.01\end{tabular} \up
& \begin{tabular}{c}100\\\hline100\end{tabular} \\ \hline

\multirow{2}{*}{$(Rust-Java)_{DD}$}
& Phi3 
\begin{tabular}{c}Baseline\\\hline KD\end{tabular}
& \begin{tabular}{c}63.77\\\hline70.28\end{tabular}
& \begin{tabular}{c}42.6\\\hline35.0\end{tabular}
& \begin{tabular}{c}51.07\\\hline46.72\end{tabular} \down
& \begin{tabular}{c}100\\\hline100\end{tabular} \\ \cline{2-6}

& Qwen-Coder 
\begin{tabular}{c}Baseline\\\hline KD\end{tabular}
& \begin{tabular}{c}72.16\\\hline70.96\end{tabular}
& \begin{tabular}{c}64.8\\\hline69.4\end{tabular}
& \begin{tabular}{c}68.28\\\hline70.17\end{tabular} \up
& \begin{tabular}{c}100\\\hline100\end{tabular} \\ \hline

\multirow{2}{*}{$(Rust-Python)_{DD}$}
& Phi3 
\begin{tabular}{c}Baseline\\\hline KD\end{tabular}
& \begin{tabular}{c}65.95\\\hline70.48\end{tabular}
& \begin{tabular}{c}37.2\\\hline32.0\end{tabular}
& \begin{tabular}{c}47.57\\\hline44.01\end{tabular} \down
& \begin{tabular}{c}100\\\hline100\end{tabular} \\ \cline{2-6}

& Qwen-Coder 
\begin{tabular}{c}Baseline\\\hline KD\end{tabular}
& \begin{tabular}{c}69.62\\\hline69.32\end{tabular}
& \begin{tabular}{c}55.0\\\hline 59.2\end{tabular}
& \begin{tabular}{c}61.45\\\hline63.86\end{tabular} \up
& \begin{tabular}{c}100\\\hline100\end{tabular} \\ \hline

\multirow{2}{*}{$(Rust-Ruby)_{DD}$}
& Phi3 
\begin{tabular}{c}Baseline\\\hline KD\end{tabular}
& \begin{tabular}{c}63.79\\\hline66.45\end{tabular}
& \begin{tabular}{c}44.4\\\hline41.6\end{tabular}
& \begin{tabular}{c}52.35\\\hline51.16\end{tabular} \down
& \begin{tabular}{c}100\\\hline100\end{tabular} \\ \cline{2-6}

& Qwen-Coder 
\begin{tabular}{c}Baseline\\\hline KD\end{tabular}
& \begin{tabular}{c}67.99\\\hline68.90\end{tabular}
& \begin{tabular}{c}61.6\\\hline61.06\end{tabular}
& \begin{tabular}{c}64.63\\\hline65.04\end{tabular} \up
& \begin{tabular}{c}100\\\hline100\end{tabular} \\ \hline

\end{tabular}}
\end{table}

We compare three approaches for ensuring complete responses: standard generation, forced conclusion, and classification-head variants (Tables~\ref{tab:combined-kd}, \ref{tab:combined-force-kd}, \ref{tab:combined-binary-head-kd}, and \ref{tab:combined-contrastive-kd}).

Standard generation (Table~\ref{tab:combined-kd}) fails to achieve reliable response coverage, with response rates often below 50\% even after KD. In contrast, both forced conclusion (Table~\ref{tab:combined-force-kd}) and classification-heads (Table \ref{tab:combined-binary-head-kd} and Table \ref{tab:combined-contrastive-kd}) approaches achieve 100\% response rate across all settings.

However, these methods differ significantly in detection performance. The forced conclusion method consistently achieves the best trade-off between completeness and accuracy. Under KD, it yields F1 scores around 70--75 across all SD and DD test sets for both models (Table~\ref{tab:combined-force-kd}). In contrast, the binary classification head (Table \ref{tab:combined-binary-head-kd}) shows substantially lower F1, particularly due to reduced recall (e.g., Phi3 recall as low as 29.8 on $(Rust$-$Python)_{DD}$), leading to F1 scores in the 40--50 range.

The contrastive head (Table \ref{tab:combined-contrastive-kd}) shows slightly better balance than the binary head but still underperforms compared to forced conclusion, especially on DD test sets. Moreover, KD provides inconsistent improvements in these classification-head settings, in contrast to its consistently positive impact when combined with forced conclusion.

A key insight is that enforcing a final answer within the generative framework is more effective than replacing it with a classification head. This suggests that the underlying reasoning capability of generative models remains valuable, provided that output completion is guaranteed.

\begin{tcolorbox}[
  colback=gray!8,
  colframe=black!70,
  boxrule=0.6pt,
  arc=2mm,
  left=1mm,
  right=1mm,
  top=1mm,
  bottom=1mm
]
\textbf{Conclusion for RQ3.} The forced conclusion method is the most effective approach for achieving complete response rates. It guarantees 100\% response coverage while maintaining significantly higher and more stable F1 scores than classification-head alternatives. When combined with knowledge distillation, it provides the best overall performance across all evaluation settings.
\end{tcolorbox}

\section{Discussion}\label{discussion-section}

This section provides a deeper interpretation of the results presented earlier and discusses the impact of different design choices on model performance. In particular, we analyze the effect of knowledge distillation, response formatting, and architectural modifications across both same-distribution (SD) and different-distribution (DD) settings. All analyses in this section are conducted on the same evaluation sets used for RQ1--RQ3 in the Results section; therefore, the discussion below focuses on interpreting the observed patterns rather than introducing new test sets or experimental settings. We focus on understanding how these factors influence performance, stability, and generalization across language pairs. We also discuss the trade-offs between different approaches in terms of effectiveness and computational cost, and highlight key observations and limitations of our study.

\subsection{Scaling F1 scores for Better Performance Intuition}

To better interpret the results and isolate the contribution of knowledge distillation, we analyze the scaled F1 scores derived from Table~\ref{tab:combined-kd} and summarized in Figure~\ref{fig:scaled-results-pure-model}. Additionally, we calculate the average F1 score across different test distributions (SD and DD) and for different models, as reported in Table~\ref{tab:final_results_improvement}. 
The scaling procedure removes the effect of response rate by treating non-interpretable outputs as failures, providing a more conservative and realistic estimate of model performance.

\begin{table}[H]
\centering
\caption{Scaled average F1 scores of results represented in Table \ref{tab:combined-kd}, where the non-interpretable outputs are considered as incorrect classification. The results are shown for both same-distribution (SD) and different-distribution (DD). }
\begin{tabular}{lcc}
\hline
\textbf{Metric} & \textbf{Phi3} & \textbf{Qwen-Coder} \\
\hline
SD (Baseline) & 21.50 & 24.66 \\
SD (KD)     & 28.42 & 51.94 \\
\hline
Improvement & 6.92 & 27.28 \\
\hline
DD (Baseline) & 19.15 & 26.49 \\
DD (KD)     & 27.99 & 49.87 \\
\hline
Improvement & 8.84  & 23.38 \\
\hline
\end{tabular}

\label{tab:final_results_improvement}
\end{table}

\begin{figure}[htbp]
\centering
\includegraphics[width=0.9\textwidth]{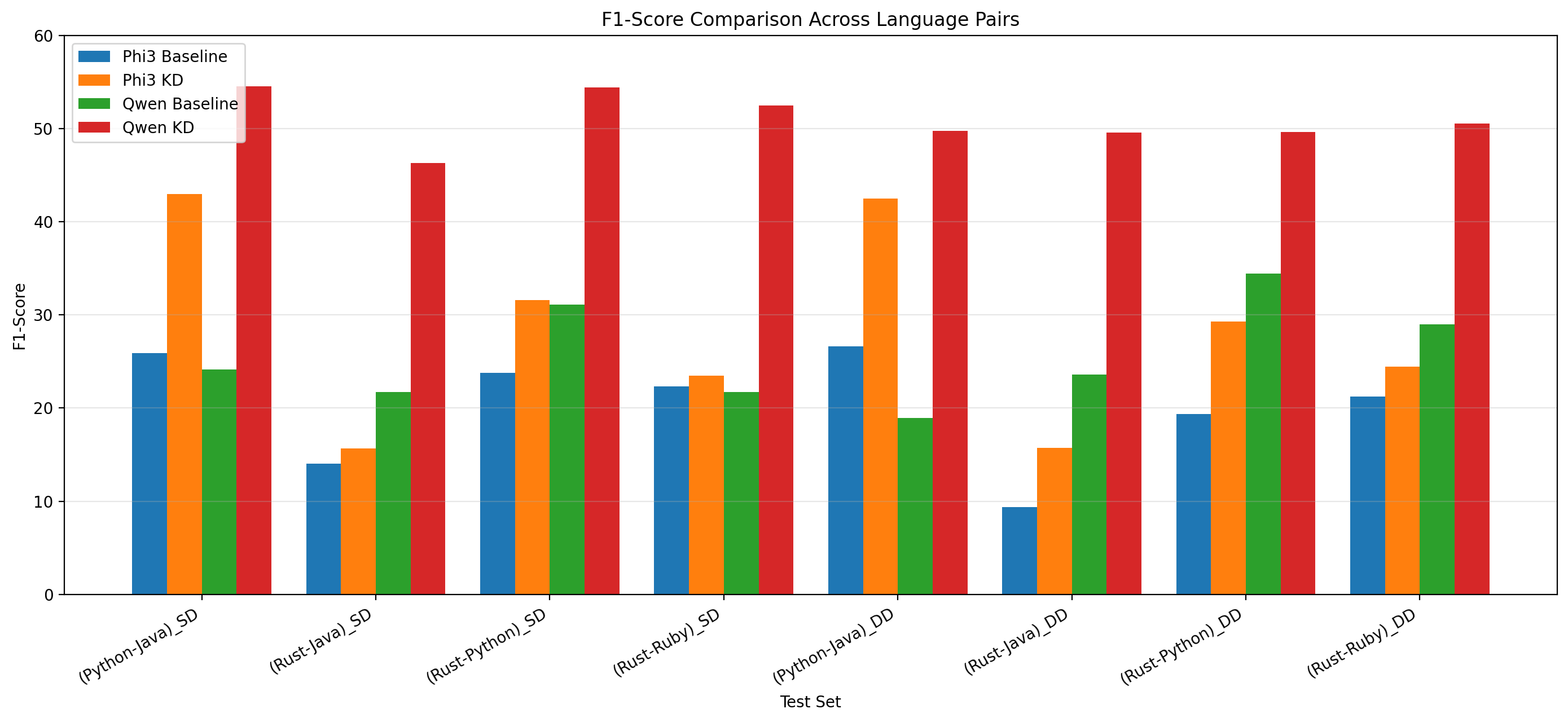}
\caption{F1 score comparison across language pairs for Phi3 and Qwen with and without knowledge distillation after scaling the results of base models.}
\label{fig:scaled-results-pure-model}
\end{figure}

Across both same-distribution and different-distribution settings, KD consistently improves performance for both models. For Phi3, the F1 score increases from 21.50 to 28.42 in SD and from 19.15 to 27.99 in DD, corresponding to gains of 6.92 and 8.84 points, respectively. For Qwen-Coder, the improvements are significantly larger, with SD increasing from 24.66 to 51.94 (+27.28) and DD from 26.49 to 49.87 (+23.38). These gains correspond to relative improvements of approximately 32--46\% for Phi3 and over 90--110\% for Qwen-Coder, highlighting a substantial disparity in how effectively the two models benefit from KD.

Averaging across SD and DD, Phi3 achieves an overall improvement of approximately 7.88 points, whereas Qwen-Coder achieves a markedly higher overall improvement of approximately 25.33 points. This consistent gap suggests that KD effectiveness is strongly dependent on the underlying model. One plausible explanation is that Qwen-Coder, being more specialized for code-related tasks, is better aligned with the teacher model and therefore more capable of capturing and retaining the distilled knowledge. In contrast, Phi3 appears more limited in leveraging the additional supervision signal provided by KD.

Comparing SD and DD settings reveals further differences in model behavior. Phi3 exhibits slightly larger gains under distribution shift (DD) than in SD, suggesting that KD may be particularly helpful for improving its robustness to unseen problem distributions. For Qwen-Coder, the largest gain is observed in the SD setting; however, the DD improvement remains substantial, increasing from 26.49 to 49.87 scaled F1. This shows that KD benefits Qwen-Coder in both in-distribution and distribution-shifted settings, even though its relative advantage is stronger in SD. Overall, these results suggest that KD improves robustness and performance across both models, but the magnitude and distribution-specific effects vary by model.

At the level of individual language pairs (Figure~\ref{fig:scaled-results-pure-model}), KD leads to consistent and substantial improvements for Qwen-Coder across all evaluated pairs, often pushing F1 scores to around or above 50 F1 score. In contrast, Phi3 demonstrates more moderate and heterogeneous gains. Certain language pairs benefit more than others, suggesting that the effectiveness of KD depends on the structural similarity and transferability between languages for this model. This variability indicates that KD does not uniformly enhance all language pairs, and that pair-specific characteristics, such as syntax similarity or dataset alignment, may influence the degree of improvement.

Comparing the two models more broadly, Qwen-Coder not only achieves higher absolute performance but also exhibits significantly greater sensitivity to KD. The improvement margin for Qwen-Coder is more than three times larger than that of Phi3 in SD and remains substantially higher in DD. This observation underscores the importance of model capacity, architectural design, and pretraining alignment in determining how effectively KD can be exploited.

Overall, several key observations emerge:
\begin{itemize}
    \item KD consistently improves performance across all settings and language pairs, confirming its effectiveness as a general enhancement technique.
    \item The magnitude of improvement is strongly model-dependent, with Qwen-Coder benefiting disproportionately more than Phi3.
    \item KD contributes to both in-distribution performance (SD) and cross-distribution generalization (DD), although the relative impact differs across models.
    \item Improvements are not uniform across language pairs, indicating that transfer effectiveness depends on task-specific and language-specific characteristics.
\end{itemize}

Despite these gains, several limitations should be noted. First, the scaling procedure penalizes all non-interpretable outputs uniformly, which may overestimate errors in borderline cases and obscure partial correctness. Second, averaging results across language pairs may mask fine-grained behaviors and pair-specific trends. Finally, the observed differences between models suggest that KD effectiveness is contingent on factors such as model architecture and pretraining, which are not explicitly controlled in this study.

Taken together, these findings demonstrate that while KD is a powerful technique for improving performance, its impact is highly conditional on the student model and task characteristics. This highlights the need for more targeted distillation strategies that account for model-specific and cross-language variations.

\subsection{Effect of forced conclusion on Model Performance}

To further understand the impact of response formatting on model performance, we analyze the effect of enforcing a forced conclusion, as illustrated in Figure~\ref{fig:f1_comparison_3_}. This setting ensures that the model always produces a definitive binary output, thereby reducing ambiguity in responses.
\begin{figure}[htbp]
\centering
\includegraphics[width=0.9\textwidth]{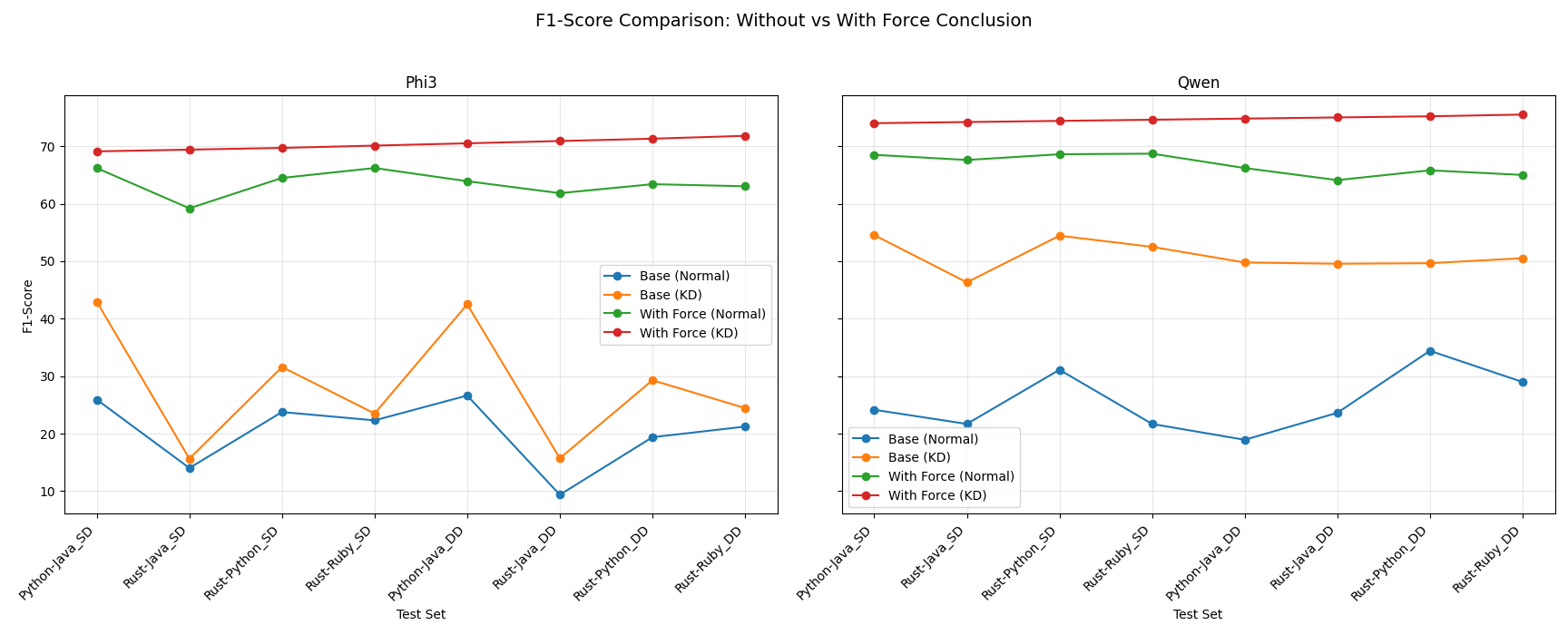}
\caption{Line chart comparing F1 scores without forced conclusion (Base model) and with forced conclusion under Baseline and KD settings.}
\label{fig:f1_comparison_3_}
\end{figure}

Across both models, enforcing a forced conclusion leads to substantial improvements in F1 scores under both Baseline and KD settings. For Phi3, the performance gains are particularly significant when comparing the base model without forced conclusion to its counterpart with forced conclusion. The baseline F1 scores (Baseline) are consistently low and highly variable across language pairs, whereas enforcing a conclusion significantly stabilizes and elevates performance, bringing scores into a much higher and more consistent range. A similar trend is observed under KD, where forced conclusion further boosts already improved scores.

For Qwen-Coder, the effect of forced conclusion is also consistently positive, although less dramatic than for Phi3. The base Qwen model already achieves relatively higher F1 scores compared to Phi3, but enforcing a conclusion still yields noticeable gains across all language pairs. In particular, the combination of KD and forced conclusion produces the highest overall performance, with F1 scores consistently exceeding those of all other configurations.

A key observation from Figure~\ref{fig:f1_comparison_3_} is that forced conclusion reduces performance variability across language pairs for both models. Without this constraint, the models occasionally produce ambiguous or non-interpretable outputs, leading to lower and more inconsistent F1 scores. By enforcing a definitive output, the models exhibit more stable behavior, suggesting that a significant portion of the observed errors in the base setting arises from response format issues rather than purely incorrect reasoning.

Comparing the interaction between KD and forced conclusion reveals complementary effects. While KD improves the model’s ability to produce correct predictions, forced conclusion ensures that these predictions are consistently expressed in an interpretable format. 

Overall, these results highlight that output formatting constraints can play a critical role in evaluating and improving model performance. In particular, the gains achieved through forced conclusion suggest that part of the performance gap observed in earlier analyses may be attributed to response ambiguity rather than inherent model limitations. This finding underscores the importance of carefully designing evaluation protocols and prompts to ensure fair and consistent assessment of model capabilities.

\subsection{Effect of Binary and Contrastive Classification Heads}

To further control the output space and eliminate ambiguity in generation, we augment the base language models with explicit classification heads. Specifically, we first train a binary classification head on top of the base model to enforce deterministic outputs (0 or 1) for code clone detection. Building on this, we introduce a contrastive classification head, motivated by prior work \cite{du2024adaccd} suggesting that contrastive objectives can improve robustness and stability, particularly for Different Distribution (DD) or unseen samples.

\begin{figure}[htbp]
\centering
\includegraphics[width=0.9\textwidth]{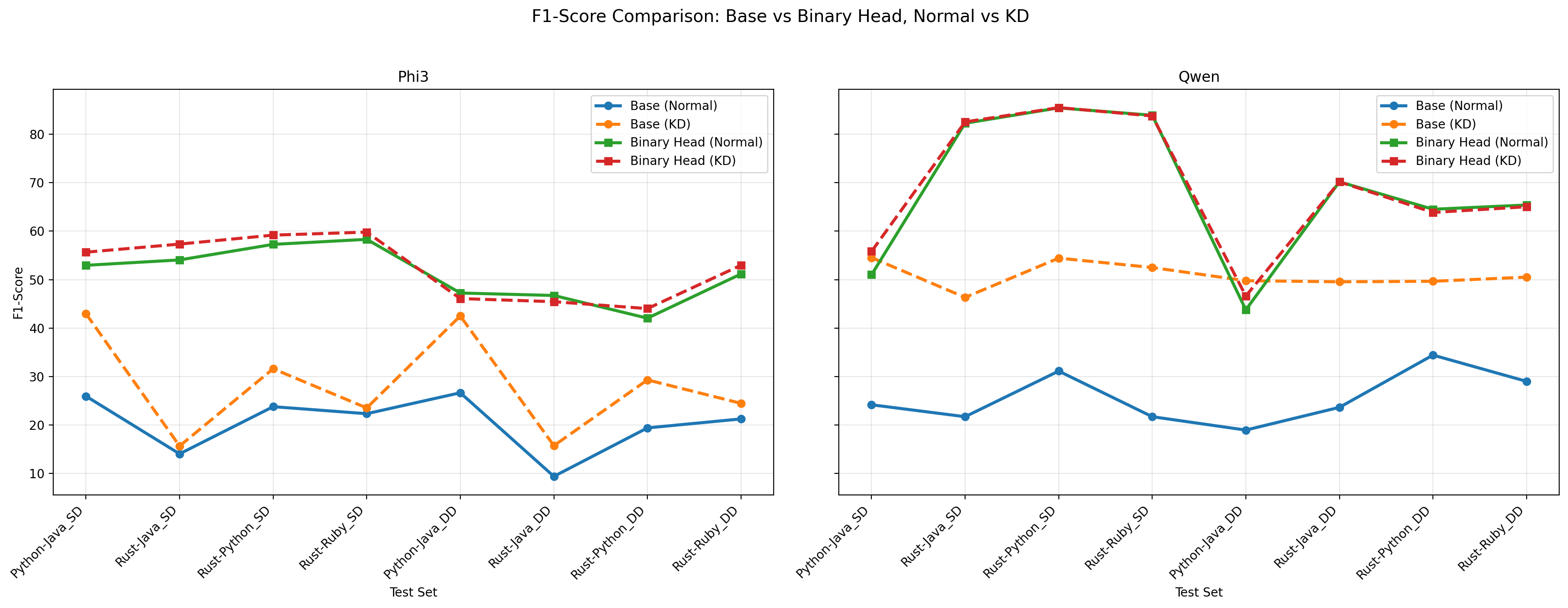}
\caption{Line chart comparing F1 scores for base models and the binary classification head under Baseline and KD settings.}
\label{fig:binary-head}
\end{figure}

\begin{figure}[htbp]
\centering
\includegraphics[width=0.9\textwidth]{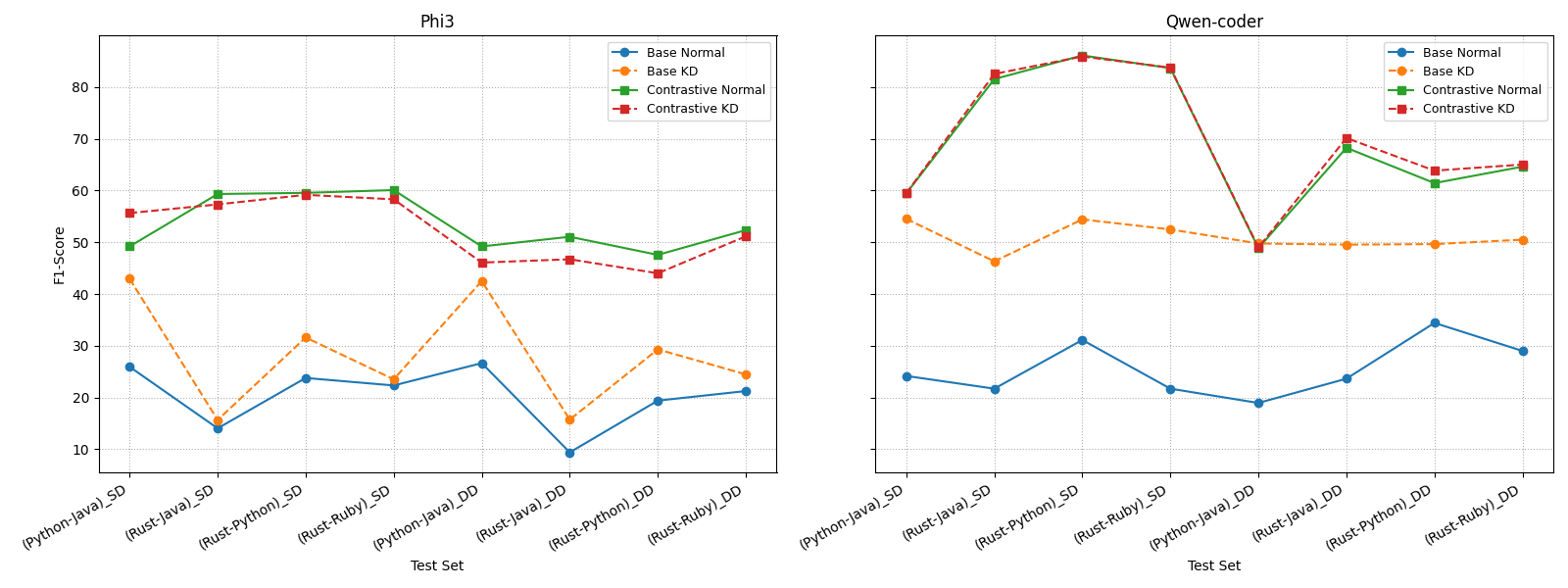}
\caption{Line chart comparing F1 scores for base models and the contrastive classification head under Baseline and KD settings.}
\label{fig:contrastive-head}
\end{figure}

As shown in Figure~\ref{fig:binary-head}, introducing a binary classification head leads to substantial improvements over the base model for both Phi3 and Qwen-Coder across all language pairs and settings. The gains are consistent in both SD and DD scenarios, confirming that constraining the output space to a discrete decision boundary significantly reduces ambiguity and improves prediction reliability. Notably, the improvements are especially noticeable in DD settings, indicating that the binary formulation already contributes to better generalization. 
However, compared to the forced conclusion approach, the classification-head results fluctuate more across language pairs and distributions.
Figure~\ref{fig:contrastive-head} compares the base models with the addition of a contrastive head. While prior work suggests that contrastive approaches lead to more robust performance, particularly for unseen problems, our results show that the contrastive head follows patterns similar to those observed with the binary classification head, with performance varying across different programming language pairs. 

That being said, although the binary classification head results in a lower F1 score for Python--Java pairs under different distribution for the Qwen model compared to the knowledge-distilled baseline, the contrastive head maintains a comparable level of performance. One possible explanation for this observation is the simplicity of the contrastive objective used in this study. It is likely that both the accuracy and robustness of the contrastive head could be improved by designing a more effective contrastive objective function.

\subsection{Comparison of Methods: Performance vs. Execution Time Trade-offs}

Figure~\ref{fig:method-performance} compares the average performance of the methods introduced in this study when applied to Phi3 and Qwen-Coder. The reported averages are computed over same-distribution, different-distribution, and overall performance. As shown in the figure, the forced conclusion approach achieves the highest overall F1 score, reaching 73.9 for Qwen-Coder and 70.3 for Phi3, outperforming both the binary and contrastive classification heads in terms of aggregated performance.

However, a more fine-grained analysis reveals that the relative effectiveness of these methods depends on the evaluation setting. In particular, for Qwen-Coder under SD, the binary classification head achieves the highest F1 score (78.0), exceeding the performance of the forced conclusion approach (72.7) on the same setting. This indicates that explicitly training a classification head can be more effective when the test distribution closely matches the training distribution. In contrast, the forced conclusion method provides more consistent improvements across both SD and DD, leading to higher overall averages despite not always achieving the best peak performance.

These observations highlight an important trade-off between peak accuracy and robustness. While classification heads (binary and contrastive) can achieve higher performance in favorable conditions (e.g., SD), the forced conclusion approach demonstrates more stable behavior across distribution shifts, which contributes to its superior overall performance.

Despite these performance gains, Table~\ref{tab:execution-time} shows that the forced conclusion approach incurs a significant computational cost. Specifically, it exhibits the highest inference time among all evaluated methods, requiring 744 minutes for Phi3 and 786 minutes for Qwen-Coder. This is nearly twice the inference time of the baseline models, which already require several hours to complete. 

In contrast, the binary and contrastive classification heads are substantially more efficient, with average inference times of only a few minutes across the evaluation settings (approximately 1.8--2.7 minutes across models). Compared with base model inference, the classification-head approaches reduce inference time by approximately 209--243$\times$ for Phi3 and 176--198$\times$ for Qwen-Coder. Compared with forced conclusion, they reduce inference time by approximately 354--413$\times$ for Phi3 and 291--328$\times$ for Qwen-Coder, as reported in Table~\ref{tab:execution-time}. The reduced computational overhead stems from the fact that classification heads transform the task into a direct prediction problem, avoiding the need for iterative or verbose generation.
Taken together, these results reveal a clear trade-off between performance and efficiency. While the forced conclusion approach yields the highest overall accuracy and more stable performance across distributions, it does so at a significantly higher computational cost. On the other hand, classification heads provide a highly efficient alternative, achieving competitive performance with dramatically lower inference time. 
These findings suggest that the choice of method should be guided by application requirements. In scenarios where computational resources and latency are critical, classification heads offer a practical and scalable solution. Conversely, when maximizing predictive performance and robustness is the primary objective, the forced conclusion approach may be preferred despite its higher cost.
\begin{figure}[htbp]
\centering
\includegraphics[width=1\textwidth]{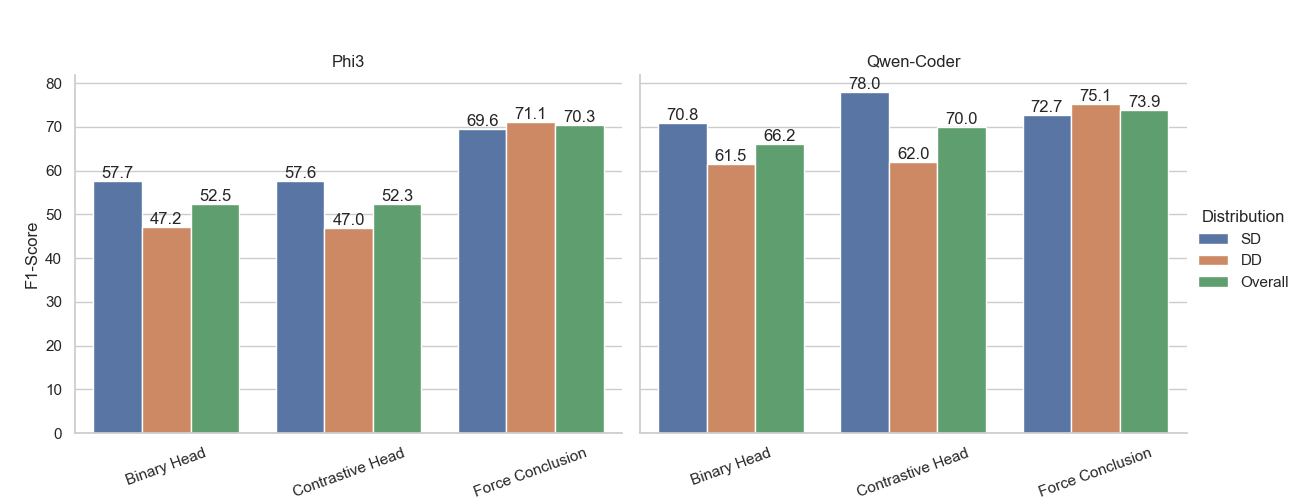}
\caption{Comparison of different methods (binary head, contrastive head, and forced conclusion) across SD and DD settings.}
\label{fig:method-performance}
\end{figure}

\begin{table}[h!]
\centering
\renewcommand{\arraystretch}{1.2}
\setlength{\tabcolsep}{6pt}
\caption{Average execution time comparison (in minutes) across different approaches for Phi3 and Qwen-Coder across all evaluation settings.}
\label{tab:execution-time}
\begin{tabular}{|l|c|c|}
\hline
\textbf{Approach} & \textbf{Phi3 (min)} & \textbf{Qwen-Coder (min)} \\ \hline

Base Model Inference & 438 & 474 \\ \hline
forced conclusion & 744 & 786 \\ \hline
Binary Classification Head & 1.8 & 2.4 \\ \hline
Contrastive Classification Head & 2.1 & 2.7 \\ \hline

\end{tabular}
\end{table}

\subsection{Implications}
The findings of this study have several implications for both practitioners and researchers working on X-CCD. First, our results show that compact open-source language models can become more practical for X-CCD when they are adapted using reasoning-oriented knowledge distillation. This is important because many real-world software engineering settings cannot rely on proprietary API-based models due to cost, privacy, latency, or reproducibility concerns. By distilling reasoning traces from a strong teacher model into smaller student models, practitioners can deploy models locally while still benefiting from the semantic reasoning ability of larger LLMs.

Second, the results suggest that response reliability is as important as predictive performance when using LLMs for binary software engineering tasks. In generation-based settings, a model may produce a partially correct or semantically meaningful explanation, but the output may still fail to map to a valid clone/non-clone label. This makes direct evaluation difficult and reduces the usefulness of the model in automated pipelines. The response rate metric introduced in this work helps capture this behavior explicitly. For practitioners, this means that evaluating only the F1 score, precision, or recall may be insufficient when the model output is generated in natural language. A model with strong reasoning ability is not necessarily useful unless its output can be consistently converted into an actionable decision.

Third, our response-stabilization methods provide different practical trade-offs. The forced conclusion approach is useful when interpretability and reasoning are important, because it allows the model to first generate a reasoning-based response and then converts that reasoning into a deterministic binary decision. This makes it suitable for settings where developers may want to inspect the model's reasoning before accepting the final prediction. However, this benefit comes with a high inference cost because the model still needs to generate long reasoning responses. In contrast, the binary and contrastive classification heads are more suitable for large-scale clone detection scenarios, where thousands of code pairs must be processed efficiently. These heads remove the need for natural-language generation during inference and reduce execution time from several hours to only a few minutes, while still producing a complete response rate.

Finally, the comparison between binary and contrastive classification heads suggests that adding a classification layer on top of an autoregressive language model is a promising direction for making LLM-based CCD systems more efficient and reliable. The binary head provides a simple and strong baseline for deterministic prediction, while the contrastive head introduces an additional representation-learning objective that encourages clone and non-clone pairs to be more separable in the embedding space. Although the contrastive head did not consistently outperform the binary head in all settings, it highlights an important research direction: future work can design more effective contrastive objectives that better capture cross-language semantic similarity. This is especially relevant for Type-4 clones, where syntactic similarity is limited and semantic alignment is essential.

Overall, these implications suggest that LLM-based X-CCD should not be treated only as a prompting problem. Instead, practical systems need to combine reasoning-oriented adaptation, reliable output stabilization, and efficient inference mechanisms. Our results show that knowledge distillation improves the semantic ability of compact models, while response-stabilization methods make their outputs more reliable and usable in real-world software engineering workflows. 
Our work can provide the building blocks for future research in this direction for X-CCD or other software engineering tasks that require reasoning and suffer from response stability. 

\section{Related Work}
\label{ltr}

\subsection{Code Clone Detection and X-CCD}
Code clone detection has been widely investigated in software engineering, with early approaches primarily focusing on syntactic similarities within source code written in the same programming language. Classical token-based, tree-based, and metric-based techniques are effective for Type-1 to Type-3 clones, but they generally struggle with semantic clones, especially when code fragments differ substantially in structure or are implemented in different programming languages \cite{baxter1998clone,kamiya2002ccfinder,jiang2007deckard}.

To address semantic clones and X-CCD, recent work has increasingly adopted learning-based approaches. Models such as CodeBERT, GraphCodeBERT, UniXcoder, and CodeT5 learn code representations from lexical tokens, syntactic structure, data-flow information, identifier semantics, and natural-language context \cite{codebert,graphcodebert,unixcoder,codet5}. By mapping code written in different programming languages into a shared embedding space, these models enable similarity-based or classification-based clone detection. More recent models and benchmarks further emphasize multilingual and cross-lingual evaluation. For example, CodeSage and CoLSBERT improve code understanding through larger pre-training corpora and contrastive objectives \cite{codesage,colsbert}, while CCT-Code introduces cross-consistency training for multilingual clone detection and code search \cite{sorokin2025cctcode}. Lin et al.~\cite{lin2026cl4d} propose CL4D, a contrastive learning framework that adapts decoder-only code generation models for code understanding tasks, including clone detection.

LLMs have also been studied for code clone detection. Prior work shows that LLMs can capture high-level semantic relationships between code fragments and may perform well on semantic clone detection within a single language \cite{dou2023}. However, recent evidence suggests that LLMs are less reliable in cross-language settings. Moumoula et al.~\cite{moumoula2025struggles} compare LLMs and embedding-based models for X-CCD and show that embedding-based approaches with lightweight classifiers can outperform prompt-based LLMs, especially on more complex problems. These findings suggest that learned representations remain important for robust X-CCD.

\subsection{Knowledge Distillation in Software Engineering}
Knowledge distillation (KD) transfers knowledge from a large teacher model to a compact student model, often reducing inference cost while preserving much of the teacher's performance \cite{hinton2015distilling}. KD has been widely studied in NLP and computer vision \cite{sanh2019distilbert,jiao2020tinybert,wang2020minilm}, and it is increasingly relevant to software engineering because code intelligence tasks often require efficient inference over large repositories. Prior SE works such as Compressor and Avatar use KD to compress code models and optimize student configurations for code intelligence tasks \cite{compressor,avatar}.

Recent work has begun to study KD more systematically for code understanding. Wang et al.~\cite{wang2025kdcode} evaluate logit-based and feature-based KD methods across multiple student architectures and teacher models on defect detection, clone detection, and exception classification. They show that KD generally improves student models over standard fine-tuning, that code-specific teachers are often more effective than general-purpose teachers, and that feature-based KD can preserve much of the teacher's performance with substantially fewer parameters. However, this study focuses on general code understanding rather than X-CCD.

\subsection{Knowledge Distillation and Response Stabilization for X-CCD}
Although KD has been studied for code understanding, its use for CCD and especially X-CCD remains limited. Existing X-CCD methods mainly focus on representation learning, graph encoders, contrastive learning, or LLM prompting \cite{moumoula2025struggles,graphcodebert,unixcoder}. These works improve clone detection accuracy, but they do not explicitly study how knowledge from a large teacher model can be transferred to an efficient student model for cross-language clone detection.

Our work is also related to response stabilization, which aims to reduce output variability across prompts, decoding paths, or repeated runs. Self-consistency improves reasoning by sampling multiple reasoning paths and aggregating final answers \cite{wang2023selfconsistency}, and multi-perspective self-consistency applies similar ideas to code generation \cite{huang2024mpsc}. IdentityChain evaluates whether code LLMs behave consistently under semantically equivalent transformations \cite{min2024identitychain}. These works show that response aggregation and consistency checking can improve reliability, but they do not study KD for X-CCD. In contrast, our work focuses on transferring stabilized teacher knowledge to compact student models for accurate and efficient cross-language clone detection.

Overall, the closest works to ours are Moumoula et al.~\cite{moumoula2025struggles}, Wang et al.~\cite{wang2025kdcode}, and Lin et al.~\cite{lin2026cl4d}. Moumoula et al. study LLMs and embedding models for X-CCD, but not KD or response stabilization. Wang et al. study KD for code understanding, including clone detection, but not X-CCD. Lin et al. improve decoder-only models for code understanding and clone detection using contrastive learning, but do not distill stabilized teacher responses into efficient X-CCD models. Our novelty lies in combining KD and response stabilization for X-CCD, enabling compact student models to learn cross-language semantic clone knowledge while reducing both inference cost and prediction variability.

\section{Threats to Validity} \label{validity}
Despite our efforts to design a rigorous experimental methodology, several factors may affect the validity and generalizability of our findings.

\subsection{Internal Validity}

One threat to internal validity arises from the sensitivity of small language models to prompt formulation. Even minor variations in prompt wording, structure, or formatting can lead to noticeably different outputs. Since our study relies on specific reasoning-oriented and task-specific prompts for both teacher querying and student inference, part of the observed performance may depend on these prompt choices rather than only on the proposed knowledge distillation and response stabilization methods. We tried to address this issue through preliminary experiments with various prompts. 

Another internal threat concerns the construction of the distilled training data. Our knowledge distillation pipeline depends on outputs generated by DeepSeek-R1, and any noise, inconsistency, or bias in the teacher responses may propagate to the student models. Although we mitigate this threat by filtering samples whose teacher-predicted labels do not agree with the ground truth, the remaining data may still reflect teacher-specific reasoning patterns and preferences.

A further threat is related to implementation and training choices. The observed results may be influenced by the selected hyperparameters, LoRA configuration, output length limits, and inference settings. Different choices for these parameters could lead to somewhat different performance trends. To reduce this threat, we followed commonly used LoRA fine-tuning practices and used a fixed configuration across models and experiments, including applying LoRA to all linear modules, using the same rank, alpha, dropout, learning rate, batch setup, validation split, and random seed. We also kept the inference procedure consistent within each experimental setting.

\subsection{External Validity}

The main threat to external validity concerns the generalizability of our findings beyond the models, datasets, and programming languages considered in this study. We evaluate two compact models, Phi3 and Qwen-Coder, both with approximately 3B parameters. Our conclusions may not transfer directly to larger or more advanced language models, which may exhibit different reasoning behavior, prompt sensitivity, and response stability.  
In addition, our experiments are conducted on code pairs derived from Project CodeNet and focus on four programming languages: Python, Java, Rust, and Ruby. Although these language pairs allow us to study both seen and unseen language settings, they do not cover the full diversity of programming languages, application domains, or software development styles encountered in practice. Competitive programming solutions may also differ substantially from industrial code in structure, scale, and documentation quality.

Finally, while we evaluate both same-distribution and different-distribution settings, these distributions are still derived from the same benchmark source. As a result, the reported generalization performance may not fully reflect behavior on other datasets, code repositories, or real-world X-CCD tasks.

\subsection{Construct Validity}

A key threat to construct validity lies in how model outputs are translated into binary clone labels. In parts of this study, regular expressions are used to extract labels from natural-language responses. The measured F1 score and response rate can vary if a different or more sophisticated extraction rule is used. As a result, the reported values may partly depend on the operationalization of label extraction rather than solely on the true clone-detection capability of the models. To address this threat, we used a deterministic analyzer consistently across all generation-based experiments and reported response rate separately from predictive performance. In addition, we introduced response-stabilization methods, including forced conclusion, binary classification head, and contrastive classification head, which avoid regular-expression-based parsing and produce a binary prediction for every input.

Construct validity is also affected by the choice of evaluation metrics. Although we introduce response rate to better capture whether a model produces a usable prediction, this metric reflects output validity rather than task correctness. Similarly, F1 score, precision, and recall capture predictive performance only after the generated outputs have been mapped to binary labels. Therefore, these measures may not fully reflect the quality of the reasoning process itself.

Finally, our operational definition of cross-language clones is based on labels derived from Project CodeNet problem-level equivalence and accepted submissions. While this is a reasonable and widely used proxy for functional equivalence, it may not capture all nuances of semantic similarity in real-world software systems.

\section{Conclusion}\label{conclusion}

In this paper, we investigated whether reasoning capabilities for X-CCD can be distilled from a strong teacher model into compact open-source language models suitable for practical deployment. We proposed a knowledge distillation pipeline based on DeepSeek-R1 and applied it to two compact student models, Phi3 and Qwen-Coder. We also introduced response stabilization methods to address a key limitation of generation-based clone detection: the inability of free-form model outputs to consistently map to valid binary labels. Our results show that knowledge distillation improves both response rate and predictive performance across the studied language pairs and evaluation settings, with particularly strong gains for Qwen-Coder. These findings suggest that teacher-generated reasoning traces provide useful supervision for improving the reliability and effectiveness of compact models in cross-language semantic clone detection.

We further showed that response stabilization is essential for making LLM-based X-CCD practically usable. The forced conclusion approach achieved the strongest overall predictive performance, while the binary classification head provided a highly efficient alternative by reducing inference time from hours to only a few minutes. The contrastive classification head showed promise, but its gains were less consistent, suggesting that more effective contrastive objectives may be needed. Overall, our findings indicate that practical X-CCD with compact open-source models depends on two complementary components: distilling task-specific reasoning from a strong teacher model and stabilizing the output space to ensure deterministic predictions. Future work can extend this direction by exploring larger multilingual datasets, improved contrastive objectives, and joint optimization of reasoning and classification-based training.


\bibliographystyle{ACM-Reference-Format}
\bibliography{reference}

@article{moumoula2025struggles,
  title={The Struggles of LLMs in Cross-Lingual Code Clone Detection},
  author={Moumoula, Micheline B{\'e}n{\'e}dicte and Kabor{\'e}, Abdoul Kader and Klein, Jacques and Bissyand{\'e}, Tegawend{\'e} F},
  journal={Proceedings of the ACM on Software Engineering},
  volume={2},
  number={FSE},
  pages={1023--1045},
  year={2025},
  publisher={ACM New York, NY, USA}
}

@article{dou2023towards,
  title={Towards understanding the capability of large language models on code clone detection: A survey},
  author={Dou, Shihan and Shan, Junjie and Jia, Haoxiang and Deng, Wenhao and Xi, Zhiheng and He, Wei and Wu, Yueming and Gui, Tao and Liu, Yang and Huang, Xuanjing},
  journal={arXiv preprint arXiv:2308.01191},
  year={2023}
}

@inproceedings{du2024adaccd,
  title={AdaCCD: Adaptive Semantic Contrasts Discovery Based Cross Lingual Adaptation for Code Clone Detection},
  author={Du, Yangkai and Ma, Tengfei and Wu, Lingfei and Zhang, Xuhong and Ji, Shouling},
  booktitle={Proceedings of the AAAI Conference on Artificial Intelligence},
  volume={38},
  number={16},
  pages={17942--17950},
  year={2024}
}

@inproceedings{khajezade2024investigating,
  title={Investigating the Efficacy of Large Language Models for Code Clone Detection},
  author={Khajezade, Mohamad and Wu, Jie JW and Fard, Fatemeh Hendijani and Rodr{\'\i}guez-P{\'e}rez, Gema and Shehata, Mohamed Sami},
  booktitle={Proceedings of the 32nd IEEE/ACM International Conference on Program Comprehension},
  pages={161--165},
  year={2024}
}

@article{lei2022deep,
  title={Deep learning application on code clone detection: A review of current knowledge},
  author={Lei, Maggie and Li, Hao and Li, Ji and Aundhkar, Namrata and Kim, Dae-Kyoo},
  journal={Journal of Systems and Software},
  volume={184},
  pages={111141},
  year={2022},
  publisher={Elsevier}
}

@article{moumoula2024large,
  title={Large Language Models for cross-language code clone detection},
  author={Moumoula, Micheline B{\'e}n{\'e}dicte and Kabore, Abdoul Kader and Klein, Jacques and Bissyande, Tegawend{\'e}},
  journal={arXiv preprint arXiv:2408.04430},
  year={2024}
}

@inproceedings{yu2019neural,
  title={Neural detection of semantic code clones via tree-based convolution},
  author={Yu, Hao and Lam, Wing and Chen, Long and Li, Ge and Xie, Tao and Wang, Qianxiang},
  booktitle={2019 IEEE/ACM 27th International Conference on Program Comprehension (ICPC)},
  pages={70--80},
  year={2019},
  organization={IEEE}
}

@inproceedings{liu2021can,
  title={Can neural clone detection generalize to unseen functionalitiesƒ},
  author={Liu, Chenyao and Lin, Zeqi and Lou, Jian-Guang and Wen, Lijie and Zhang, Dongmei},
  booktitle={2021 36th IEEE/ACM International Conference on Automated Software Engineering (ASE)},
  pages={617--629},
  year={2021},
  organization={IEEE}
}

@inproceedings{sonnekalb2022generalizability,
  title={Generalizability of code clone detection on codebert},
  author={Sonnekalb, Tim and Gruner, Bernd and Brust, Clemens-Alexander and M{\"a}der, Patrick},
  booktitle={Proceedings of the 37th IEEE/ACM International Conference on Automated Software Engineering},
  pages={1--3},
  year={2022}
}

@inproceedings{tao2022c4,
  title={C4: Contrastive cross-language code clone detection},
  author={Tao, Chenning and Zhan, Qi and Hu, Xing and Xia, Xin},
  booktitle={Proceedings of the 30th IEEE/ACM International Conference on Program Comprehension},
  pages={413--424},
  year={2022}
}

@article{zhang2021survey,
  title={A survey of software clone detection from security perspective},
  author={Zhang, Haibo and Sakurai, Kouichi},
  journal={IEEE Access},
  volume={9},
  pages={48157--48173},
  year={2021},
  publisher={IEEE}
}

@article{gunasekar2023textbooks,
  title={Textbooks are all you need},
  author={Gunasekar, Suriya and Zhang, Yi and Aneja, Jyoti and Mendes, Caio C{\'e}sar Teodoro and Del Giorno, Allie and Gopi, Sivakanth and Javaheripi, Mojan and Kauffmann, Piero and de Rosa, Gustavo and Saarikivi, Olli and others},
  journal={arXiv preprint arXiv:2306.11644},
  year={2023}
}

@article{alshabib2025systematic,
  title={A systematic literature review on cross-language source code clone detection},
  author={Alshabib, Asra Sulaiman and Mahmood, Sajjad and Alshayeb, Mohammad},
  journal={Computer Science Review},
  volume={58},
  pages={100786},
  year={2025},
  publisher={Elsevier}
}

@article{alazba2024cort,
  title={Cort: transformer-based code representations with self-supervision by predicting reserved words for code smell detection},
  author={Alazba, Amal and Aljamaan, Hamoud and Alshayeb, Mohammad},
  journal={Empirical Software Engineering},
  volume={29},
  number={3},
  pages={59},
  year={2024},
  publisher={Springer}
}

@article{mohammed2022search,
  title={A search-based approach for detecting circular dependency bad smell in goal-oriented models},
  author={Mohammed, Mawal A and Alshayeb, Mohammad and Hassine, Jameleddine},
  journal={Software and Systems Modeling},
  volume={21},
  number={5},
  pages={2007--2037},
  year={2022},
  publisher={Springer}
}

@article{wang2025empirical,
  title={An Empirical Study of Knowledge Distillation for Code Understanding Tasks},
  author={Wang, Ruiqi and Yang, Zezhou and Gao, Cuiyun and Xia, Xin and Liao, Qing},
  journal={arXiv preprint arXiv:2508.15423},
  year={2025}
}

@inproceedings{d2025compression,
  title={On the compression of language models for code: An empirical study on codebert},
  author={d'Aloisio, Giordano and Traini, Luca and Sarro, Federica and Di Marco, Antinisca},
  booktitle={2025 IEEE International Conference on Software Analysis, Evolution and Reengineering (SANER)},
  pages={12--23},
  year={2025},
  organization={IEEE}
}

@article{li2025symmetry,
  title={Symmetry-Aware Code Generation: Distilling Pseudocode Reasoning for Lightweight Deployment of Large Language Models},
  author={Li, Yonglin and Gu, Shanzhi and Geng, Mingyang},
  journal={Symmetry},
  volume={17},
  number={8},
  pages={1325},
  year={2025},
  publisher={MDPI}
}

@inproceedings{wang-etal-2023-towards,
    title = "Towards Understanding Chain-of-Thought Prompting: An Empirical Study of What Matters",
    author = "Wang, Boshi  and
      Min, Sewon  and
      Deng, Xiang  and
      Shen, Jiaming  and
      Wu, You  and
      Zettlemoyer, Luke  and
      Sun, Huan",
    editor = "Rogers, Anna  and
      Boyd-Graber, Jordan  and
      Okazaki, Naoaki",
    booktitle = "Proceedings of the 61st Annual Meeting of the Association for Computational Linguistics (Volume 1: Long Papers)",
    month = jul,
    year = "2023",
    address = "Toronto, Canada",
    publisher = "Association for Computational Linguistics",
    url = "https://aclanthology.org/2023.acl-long.153/",
    doi = "10.18653/v1/2023.acl-long.153",
    pages = "2717--2739"
}

@article{xiang2025towards,
  title={Towards system 2 reasoning in llms: Learning how to think with meta chain-of-thought},
  author={Xiang, Violet and Snell, Charlie and Gandhi, Kanishk and Albalak, Alon and Singh, Anikait and Blagden, Chase and Phung, Duy and Rafailov, Rafael and Lile, Nathan and Mahan, Dakota and others},
  journal={arXiv preprint arXiv:2501.04682},
  year={2025}
}

@article{haider2024phi,
  title={Phi-3 safety post-training: Aligning language models with a" break-fix" cycle},
  author={Haider, Emman and Perez-Becker, Daniel and Portet, Thomas and Madan, Piyush and Garg, Amit and Ashfaq, Atabak and Majercak, David and Wen, Wen and Kim, Dongwoo and Yang, Ziyi and others},
  journal={arXiv preprint arXiv:2407.13833},
  year={2024}
}

@article{puri2021codenet,
  title={Codenet: A large-scale ai for code dataset for learning a diversity of coding tasks},
  author={Puri, Ruchir and Kung, David S and Janssen, Geert and Zhang, Wei and Domeniconi, Giacomo and Zolotov, Vladimir and Dolby, Julian and Chen, Jie and Choudhury, Mihir and Decker, Lindsey and others},
  journal={arXiv preprint arXiv:2105.12655},
  year={2021}
}

@article{yahya2022cross,
  title={Cross-language source code clone detection using deep learning with infercode},
  author={Yahya, Mohammad A and Kim, Dae-Kyoo},
  journal={arXiv preprint arXiv:2205.04913},
  year={2022}
}

@article{guo2025deepseek,
  title={Deepseek-r1: Incentivizing reasoning capability in llms via reinforcement learning},
  author={Guo, Daya and Yang, Dejian and Zhang, Haowei and Song, Junxiao and Wang, Peiyi and Zhu, Qihao and Xu, Runxin and Zhang, Ruoyu and Ma, Shirong and Bi, Xiao and others},
  journal={arXiv preprint arXiv:2501.12948},
  year={2025}
}

@article{zubkov2022evaluation,
  title={Evaluation of contrastive learning with various code representations for code clone detection},
  author={Zubkov, Maksim and Spirin, Egor and Bogomolov, Egor and Bryksin, Timofey},
  journal={arXiv preprint arXiv:2206.08726},
  year={2022}
}

@inproceedings{yang2025llm2,
  title={Llm2: Let large language models harness system 2 reasoning},
  author={Yang, Cheng and Shi, Chufan and Li, Siheng and Shui, Bo and Yang, Yujiu and Lam, Wai},
  booktitle={Proceedings of the 2025 Conference of the Nations of the Americas Chapter of the Association for Computational Linguistics: Human Language Technologies (Volume 2: Short Papers)},
  pages={168--177},
  year={2025}
}

@inproceedings{bucila2006model,
  title={Model Compression},
  author={Bucil{\u{a}}, Cristian and Caruana, Rich and Niculescu-Mizil, Alexandru},
  booktitle={Proceedings of the 12th ACM SIGKDD International Conference on Knowledge Discovery and Data Mining},
  pages={535--541},
  year={2006},
  publisher={ACM},
  doi={10.1145/1150402.1150464}
}

@article{hinton2015distilling,
  title={Distilling the Knowledge in a Neural Network},
  author={Hinton, Geoffrey and Vinyals, Oriol and Dean, Jeff},
  journal={arXiv preprint arXiv:1503.02531},
  year={2015}
}

@inproceedings{hu2022lora,
  title={LoRA: Low-Rank Adaptation of Large Language Models},
  author={Hu, Edward J. and Shen, Yelong and Wallis, Phillip and Allen-Zhu, Zeyuan and Li, Yuanzhi and Wang, Shean and Wang, Lu and Chen, Weizhu},
  booktitle={International Conference on Learning Representations},
  year={2022}
}

@article{ouyang2022training,
  title={Training Language Models to Follow Instructions with Human Feedback},
  author={Ouyang, Long and Wu, Jeff and Jiang, Xu and Almeida, Diogo and Wainwright, Carroll L. and Mishkin, Pamela and Zhang, Chong and Agarwal, Sandhini and Slama, Katarina and Ray, Alex and Schulman, John and Hilton, Jacob and Kelton, Fraser and Miller, Luke and Simens, Maddie and Askell, Amanda and Welinder, Peter and Christiano, Paul and Leike, Jan and Lowe, Ryan},
  journal={Advances in Neural Information Processing Systems},
  volume={35},
  pages={27730--27744},
  year={2022}
}

@article{wei2021finetuned,
  title={Finetuned Language Models Are Zero-Shot Learners},
  author={Wei, Jason and Bosma, Maarten and Zhao, Vincent Y. and Guu, Kelvin and Yu, Adams Wei and Lester, Brian and Du, Nan and Dai, Andrew M. and Le, Quoc V.},
  journal={arXiv preprint arXiv:2109.01652},
  year={2021}
}

@misc{taori2023alpaca,
  title={Alpaca: A Strong, Replicable Instruction-Following Model},
  author={Taori, Rohan and Gulrajani, Ishaan and Zhang, Tianyi and Dubois, Yann and Li, Xuechen and Guestrin, Carlos and Liang, Percy and Hashimoto, Tatsunori B.},
  year={2023},
  howpublished={\url{https://crfm.stanford.edu/2023/03/13/alpaca.html}}
}

@article{zhang2023llamaadapter,
  title={LLaMA-Adapter: Efficient Fine-tuning of Language Models with Zero-init Attention},
  author={Zhang, Renrui and Han, Jiaming and Liu, Chris and Gao, Peng and Zhou, Aojun and Hu, Xiangfei and Yan, Shilin and Lu, Pan and Li, Hongsheng and Qiao, Yu},
  journal={arXiv preprint arXiv:2303.16199},
  year={2023}
}

@article{guo2025deepseekr1,
  title={DeepSeek-R1: Incentivizing Reasoning Capability in LLMs via Reinforcement Learning},
  author={Guo, Daya and Yang, Dejian and Zhang, Haowei and Song, Junxiao and Zhang, Ruoyu and Xu, Runxin and Zhu, Qihao and Ma, Shirong and Wang, Peiyi and Bi, Xiao and others},
  journal={arXiv preprint arXiv:2501.12948},
  year={2025}
}

@article{abdin2024phi3,
  title={Phi-3 Technical Report: A Highly Capable Language Model Locally on Your Phone},
  author={Abdin, Marah and Aneja, Jyoti and Awadalla, Hany and Awadallah, Ahmed and Awan, Ammar Ahmad and Bach, Nguyen and Bahree, Amit and Bakhtiari, Arash and Bao, Jianmin and Behl, Harkirat and others},
  journal={arXiv preprint arXiv:2404.14219},
  year={2024}
}

@article{hui2024qwen25coder,
  title={Qwen2.5-Coder Technical Report},
  author={Hui, Binyuan and Yang, Jian and Cui, Zeyu and Yang, Jiaxi and Liu, Dayiheng and Zhang, Lei and Liu, Tianyu and Zhang, Jiajun and Yu, Bowen and Lu, Keming and others},
  journal={arXiv preprint arXiv:2409.12186},
  year={2024}
}

@inproceedings{sorokin2025cctcode,
  title     = {{CCT}-Code: Cross-Consistency Training for Multilingual Clone Detection and Code Search},
  author    = {Sorokin, Nikita and Anton, Tikhonov and Abulkhanov, Dmitry and Sedykh, Ivan and Piontkovskaya, Irina and Malykh, Valentin},
  booktitle = {Proceedings of the 2025 Conference of the Nations of the Americas Chapter of the Association for Computational Linguistics: Human Language Technologies (Volume 4: Student Research Workshop)},
  pages     = {178--185},
  year      = {2025},
  month     = apr,
  address   = {Albuquerque, USA},
  publisher = {Association for Computational Linguistics},
  doi       = {10.18653/v1/2025.naacl-srw.17},
  url       = {https://aclanthology.org/2025.naacl-srw.17/}
}

@article{lin2026cl4d,
  author        = {Lin, Jiayi and Wang, Yanlin and Yang, Yibiao and Zhang, Lei and Xie, Yutao},
  title         = {Towards Better Code Understanding in Decoder-Only Models with Contrastive Learning},
  journal       = {arXiv preprint arXiv:2406.12326},
  year          = {2026},
  eprint        = {2406.12326},
  archivePrefix = {arXiv},
  primaryClass  = {cs.SE},
  url           = {https://arxiv.org/abs/2406.12326}
}

@article{wang2025kdcode,
  author        = {Wang, Ruiqi and Yang, Zezhou and Gao, Cuiyun and Xia, Xin and Liao, Qing},
  title         = {An Empirical Study of Knowledge Distillation for Code Understanding Tasks},
  journal       = {arXiv preprint arXiv:2508.15423},
  year          = {2025},
  eprint        = {2508.15423},
  archivePrefix = {arXiv},
  primaryClass  = {cs.SE},
  url           = {https://arxiv.org/abs/2508.15423}
}

@inproceedings{wang2023selfconsistency,
  author    = {Wang, Xuezhi and Wei, Jason and Schuurmans, Dale and Le, Quoc V. and Chi, Ed H. and Narang, Sharan and Chowdhery, Aakanksha and Zhou, Denny},
  title     = {Self-Consistency Improves Chain of Thought Reasoning in Language Models},
  booktitle = {Proceedings of the 11th International Conference on Learning Representations},
  year      = {2022},
  series    = {ICLR},
  url       = {https://openreview.net/forum?id=1PL1NIMMrw}
}

@inproceedings{huang2024mpsc,
  title     = {Enhancing Large Language Models in Coding Through Multi-Perspective Self-Consistency},
  author    = {Huang, Baizhou and Lu, Shuai and Wan, Xiaojun and Duan, Nan},
  booktitle = {Proceedings of the 62nd Annual Meeting of the Association for Computational Linguistics (Volume 1: Long Papers)},
  pages     = {1429--1450},
  year      = {2024},
  month     = aug,
  address   = {Bangkok, Thailand},
  publisher = {Association for Computational Linguistics},
  doi       = {10.18653/v1/2024.acl-long.78},
  url       = {https://aclanthology.org/2024.acl-long.78/}
}

@inproceedings{min2024identitychain,
  author    = {Min, Marcus J. and Ding, Yangruibo and Buratti, Luca and Pujar, Saurabh and Kaiser, Gail and Jana, Suman and Ray, Baishakhi},
  title     = {Beyond Accuracy: Evaluating Self-Consistency of Code Large Language Models with {IdentityChain}},
  booktitle = {Proceedings of the 12th International Conference on Learning Representations},
  year      = {2023},
  series    = {ICLR},
  url       = {https://openreview.net/forum?id=caW7LdAALh}
}

@inproceedings{baxter1998clone,
  author    = {Baxter, Ira D. and Yahin, Andrew and Moura, Leonardo and Sant'Anna, Marcelo and Bier, Lorraine},
  title     = {Clone Detection Using Abstract Syntax Trees},
  booktitle = {Proceedings of the International Conference on Software Maintenance},
  series    = {ICSM},
  pages     = {368--377},
  year      = {1998},
  publisher = {IEEE Computer Society},
  doi       = {10.1109/ICSM.1998.738528}
}

@inproceedings{kamiya2002ccfinder,
  author    = {Kamiya, Toshihiro and Kusumoto, Shinji and Inoue, Katsuro},
  title     = {{CCFinder}: A Multilinguistic Token-Based Code Clone Detection System for Large Scale Source Code},
  booktitle = {IEEE Transactions on Software Engineering},
  volume    = {28},
  number    = {7},
  pages     = {654--670},
  year      = {2002},
  publisher = {IEEE},
  doi       = {10.1109/TSE.2002.1019480}
}

@inproceedings{jiang2007deckard,
  author    = {Jiang, Lingxiao and Misherghi, Ghassan and Su, Zhendong and Glondu, Stephane},
  title     = {{DECKARD}: Scalable and Accurate Tree-Based Detection of Code Clones},
  booktitle = {Proceedings of the 29th International Conference on Software Engineering},
  series    = {ICSE},
  pages     = {96--105},
  year      = {2007},
  publisher = {IEEE Computer Society},
  doi       = {10.1109/ICSE.2007.30}
}

@inproceedings{codebert,
  author    = {Feng, Zhangyin and Guo, Daya and Tang, Duyu and Duan, Nan and Feng, Xiaocheng and Gong, Ming and Shou, Linjun and Qin, Bing and Liu, Ting and Jiang, Daxin and Zhou, Ming},
  title     = {{CodeBERT}: A Pre-Trained Model for Programming and Natural Languages},
  booktitle = {Findings of the Association for Computational Linguistics: EMNLP 2020},
  pages     = {1536--1547},
  year      = {2020},
  publisher = {Association for Computational Linguistics},
  doi       = {10.18653/v1/2020.findings-emnlp.139},
  url       = {https://aclanthology.org/2020.findings-emnlp.139/}
}

@inproceedings{graphcodebert,
  author    = {Guo, Daya and Ren, Shuo and Lu, Shuai and Feng, Zhangyin and Tang, Duyu and Liu, Shujie and Zhou, Long and Duan, Nan and Svyatkovskiy, Alexey and Fu, Shengyu and Tufano, Michele and Deng, Shao Kun and Clement, Colin and Drain, Dawn and Sundaresan, Neel and Yin, Jian and Jiang, Daxin and Zhou, Ming},
  title     = {{GraphCodeBERT}: Pre-training Code Representations with Data Flow},
  booktitle = {Proceedings of the 9th International Conference on Learning Representations},
  series    = {ICLR},
  year      = {2021},
  url       = {https://openreview.net/forum?id=jLoC4ez43PZ}
}

@inproceedings{unixcoder,
  author    = {Guo, Daya and Lu, Shuai and Duan, Nan and Wang, Yanlin and Zhou, Ming and Yin, Jian},
  title     = {{UniXcoder}: Unified Cross-Modal Pre-training for Code Representation},
  booktitle = {Proceedings of the 60th Annual Meeting of the Association for Computational Linguistics},
  series    = {ACL},
  pages     = {7212--7225},
  year      = {2022},
  publisher = {Association for Computational Linguistics},
  doi       = {10.18653/v1/2022.acl-long.499},
  url       = {https://aclanthology.org/2022.acl-long.499/}
}

@inproceedings{codet5,
  author    = {Wang, Yue and Wang, Weishi and Joty, Shafiq and Hoi, Steven C. H.},
  title     = {{CodeT5}: Identifier-aware Unified Pre-trained Encoder-Decoder Models for Code Understanding and Generation},
  booktitle = {Proceedings of the 2021 Conference on Empirical Methods in Natural Language Processing},
  series    = {EMNLP},
  pages     = {8696--8708},
  year      = {2021},
  publisher = {Association for Computational Linguistics},
  doi       = {10.18653/v1/2021.emnlp-main.685},
  url       = {https://aclanthology.org/2021.emnlp-main.685/}
}

@inproceedings{codesage,
  author    = {Zhang, Dejiao and Ahmad, Wasi Uddin and Tan, Ming and Ding, Hantian and Nallapati, Ramesh and Roth, Dan and Ma, Xiaofei and Xiang, Bing},
  title     = {{CodeSage}: Code Representation Learning at Scale},
  booktitle = {Proceedings of the 12th International Conference on Learning Representations},
  series    = {ICLR},
  year      = {2024},
  url       = {https://openreview.net/forum?id=vfzRRjumpX}
}

@article{colsbert,
  author        = {Lin, Jiayi and Dong, Hande and Xie, Yutao and Zhang, Lei},
  title         = {Scaling Laws Behind Code Understanding Model},
  journal       = {arXiv preprint arXiv:2402.12813},
  year          = {2024},
  eprint        = {2402.12813},
  archivePrefix = {arXiv},
  primaryClass  = {cs.SE},
  url           = {https://arxiv.org/abs/2402.12813}
}

@article{dou2023,
  author        = {Dou, Shihan and Shan, Junjie and Jia, Haoxiang and Deng, Wenhao and Xi, Zhiheng and He, Wei and Wu, Yueming and Gui, Tao and Liu, Yang and Huang, Xuanjing},
  title         = {Towards Understanding the Capability of Large Language Models on Code Clone Detection: A Survey},
  journal       = {arXiv preprint arXiv:2308.01191},
  year          = {2023},
  eprint        = {2308.01191},
  archivePrefix = {arXiv},
  primaryClass  = {cs.SE},
  url           = {https://arxiv.org/abs/2308.01191}
}

@article{sanh2019distilbert,
  author        = {Sanh, Victor and Debut, Lysandre and Chaumond, Julien and Wolf, Thomas},
  title         = {{DistilBERT}, a Distilled Version of {BERT}: Smaller, Faster, Cheaper and Lighter},
  journal       = {arXiv preprint arXiv:1910.01108},
  year          = {2019},
  eprint        = {1910.01108},
  archivePrefix = {arXiv},
  primaryClass  = {cs.CL},
  url           = {https://arxiv.org/abs/1910.01108}
}

@inproceedings{jiao2020tinybert,
  author    = {Jiao, Xiaoqi and Yin, Yichun and Shang, Lifeng and Jiang, Xin and Chen, Xiao and Li, Linlin and Wang, Fang and Liu, Qun},
  title     = {{TinyBERT}: Distilling {BERT} for Natural Language Understanding},
  booktitle = {Findings of the Association for Computational Linguistics: EMNLP 2020},
  pages     = {4163--4174},
  year      = {2020},
  publisher = {Association for Computational Linguistics},
  doi       = {10.18653/v1/2020.findings-emnlp.372},
  url       = {https://aclanthology.org/2020.findings-emnlp.372/}
}

@inproceedings{wang2020minilm,
  author    = {Wang, Wenhui and Wei, Furu and Dong, Li and Bao, Hangbo and Yang, Nan and Zhou, Ming},
  title     = {{MiniLM}: Deep Self-Attention Distillation for Task-Agnostic Compression of Pre-Trained Transformers},
  booktitle = {Advances in Neural Information Processing Systems},
  volume    = {33},
  pages     = {5776--5788},
  year      = {2020},
  url       = {https://proceedings.neurips.cc/paper/2020/hash/3f5ee243547dee91fbd053c1c4a845aa-Abstract.html}
}

@inproceedings{compressor,
  author    = {Shi, Jieke and Yang, Zhou and Xu, Bowen and Liu, Hong Jin Kang and He, Hongwei and Lu, David Lo and Lo, David},
  title     = {Compressing Pre-trained Models of Code into 3 {MB}},
  booktitle = {Proceedings of the 37th IEEE/ACM International Conference on Automated Software Engineering},
  series    = {ASE '22},
  year      = {2022},
  publisher = {Association for Computing Machinery},
  doi       = {10.1145/3551349.3556964},
  url       = {https://doi.org/10.1145/3551349.3556964}
}

@inproceedings{avatar,
  author    = {Shi, Jieke and Yang, Zhou and Xu, Bowen and Kang, Hong Jin and He, Hongwei and Lu, David Lo and Lo, David},
  title     = {Greening Large Language Models of Code},
  booktitle = {Proceedings of the 46th IEEE/ACM International Conference on Software Engineering: Software Engineering in Practice},
  series    = {ICSE-SEIP '24},
  pages     = {142--153},
  year      = {2024},
  publisher = {Association for Computing Machinery},
  doi       = {10.1145/3639475.3640097},
  url       = {https://doi.org/10.1145/3639475.3640097}
}

\appendix

\end{document}